\documentclass[10pt,twocolumn,letterpaper]{article}
\usepackage{wacv}
\usepackage{times}
\usepackage{epsfig}
\usepackage{graphicx}
\usepackage{amsmath}
\usepackage{amssymb}
\usepackage{booktabs}
\usepackage{bm}

\usepackage{comment}
\usepackage{multirow}
\usepackage{float}
\usepackage{subfig}
\usepackage{math}
\usepackage{tabularx}
\usepackage{makecell}

\newcommand{\steph}[1]{\textcolor{blue}{Steph: #1}}

\newcommand{\mbf}[1]{\mathbf{#1}}
\newcommand{\vx}{\mbf{x}}
\newcommand{\vz}{\mbf{z}}
\newcommand{\vy}{\mbf{y}}

\newcommand{\mpred}{\mbf{p}}

\newcommand{\ths}{\textsuperscript{th}\;}

\newcommand{\data}{\mathcal{D}}
\newcommand{\source}{\mathtt{S}}
\newcommand{\target}{\mathtt{T}}

\newcommand{\tsource}{\ensuremath{^{\source}}}
\newcommand{\ttarget}[1]{\ensuremath{^{\target_{#1}}}}

\newcommand{\ours}{CoaST\xspace}
\newcommand{\tcite}[1]{~\cite{#1}}

\def\wacvPaperID{752} 

\wacvalgorithmstrack   

\wacvfinalcopy 

\ifwacvfinal
\usepackage[breaklinks=true,bookmarks=false]{hyperref}
\else
\usepackage[pagebackref=true,breaklinks=true,colorlinks,bookmarks=false]{hyperref}
\fi

\begin{document}

\title{Cooperative Self-Training for Multi-Target Adaptive Semantic Segmentation}

\author{Yangsong Zhang$^{1,3}$, Subhankar Roy$^{2,4}$, Hongtao Lu$^{3}$, Elisa Ricci$^{2,4}$, Stéphane Lathuilière$^{1}$   \\
$^{1}$ LTCI, Télécom-Paris, Intitute Polytechnique de Paris \hspace{1cm} $^{2}$ University of Trento, Trento, Italy\\
$^{3}$ Shanghai Jiao Tong University, Shanghai, China \hspace{1cm} $^{4}$  Fondazione Bruno Kessler, Trento, Italy\\
{\tt\small yangsong.zhang.zys@gmail.com}
}

\maketitle
\thispagestyle{empty}
\pagestyle{empty}

\begin{abstract}
In this work we address multi-target domain adaptation (MTDA) in semantic segmentation, which consists in adapting a single model from an annotated source dataset to multiple unannotated target datasets that differ in their underlying data distributions. To address MTDA, we propose a self-training strategy that employs pseudo-labels to induce cooperation among multiple domain-specific classifiers. We employ feature stylization as an efficient way to generate image views that forms an integral part of self-training. Additionally, to prevent the network from overfitting to noisy pseudo-labels, we devise a rectification strategy that leverages the predictions from different classifiers to estimate the quality of pseudo-labels. Our extensive experiments on numerous settings, based on four different semantic segmentation datasets, validates the effectiveness of the proposed self-training strategy and shows that our method outperforms state-of-the-art MTDA approaches. Code available at: \href{https://github.com/Mael-zys/CoaST}{https://github.com/Mael-zys/CoaST}.

\end{abstract}

\section{Introduction}
\label{sec:intro}

Semantic segmentation is a key task in computer vision that consists in learning to predict semantic labels for image pixels. Given its importance in many real world applications, segmentation is widely studied and significant progress has been made \cite{badrinarayanan2017segnet,chen2017deeplab,chen2017rethinking} in the supervised regime. Much of the recent success can be attributed to the availability of large, curated, and annotated datasets \cite{cordts2016cityscapes,neuhold2017mapillary,zhou2017scene}. As obtaining labeled data in semantic segmentation is costly and tedious, pre-trained models are often deployed in test environments without fine-tuning. Unfortunately, these models fail when the test samples are drawn from a distribution which is different from the training distribution. This phenomenon is known as the \textit{domain shift} \cite{torralba2011unbiased} problem. To mitigate the domain-shift 
between the training (\textit{source}) and test (\textit{target}) distributions, \textit{Unsupervised Domain Adaptation} (UDA) methods \cite{csurka2021unsupervised} have been proposed. 

\begin{figure}[t]%
    \centering
    \subfloat[\centering Cooperative Self-training]{\includegraphics[width=0.99\columnwidth]{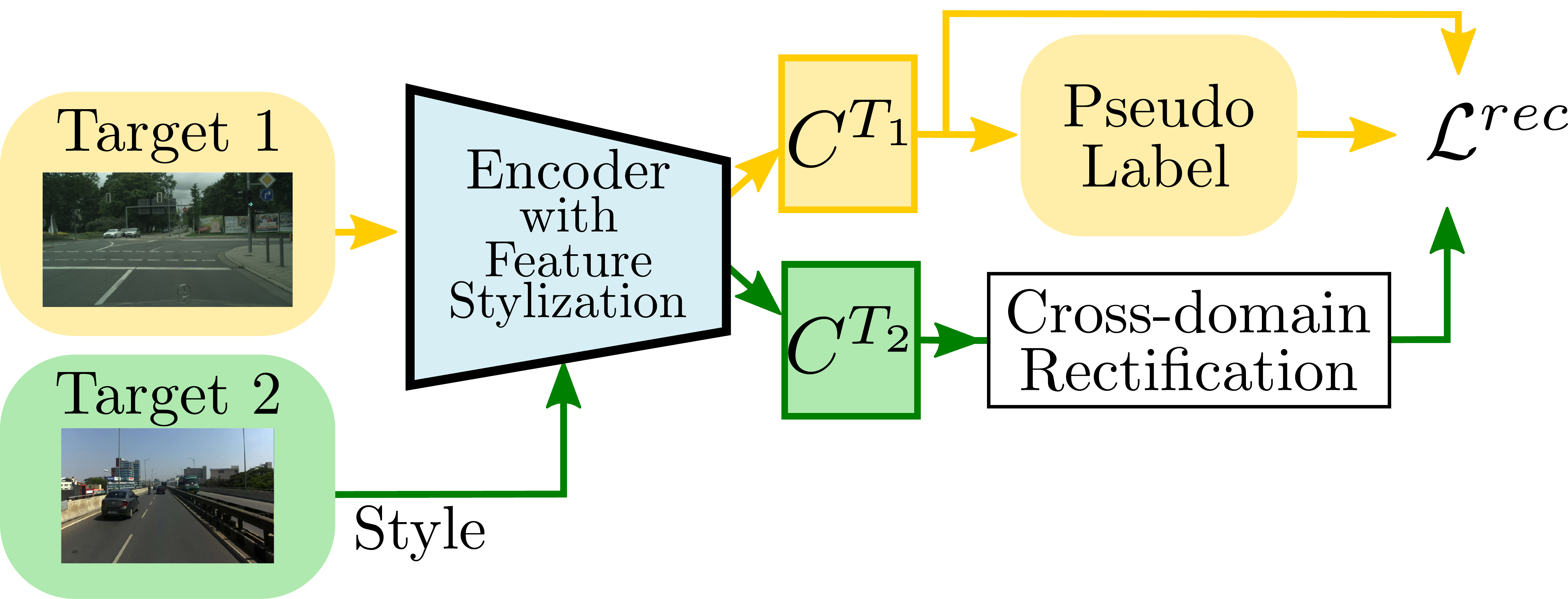}}\\
    \vspace{-3mm}
    \subfloat[\centering Cooperative
    Rectification]{\includegraphics[width=0.99\columnwidth]{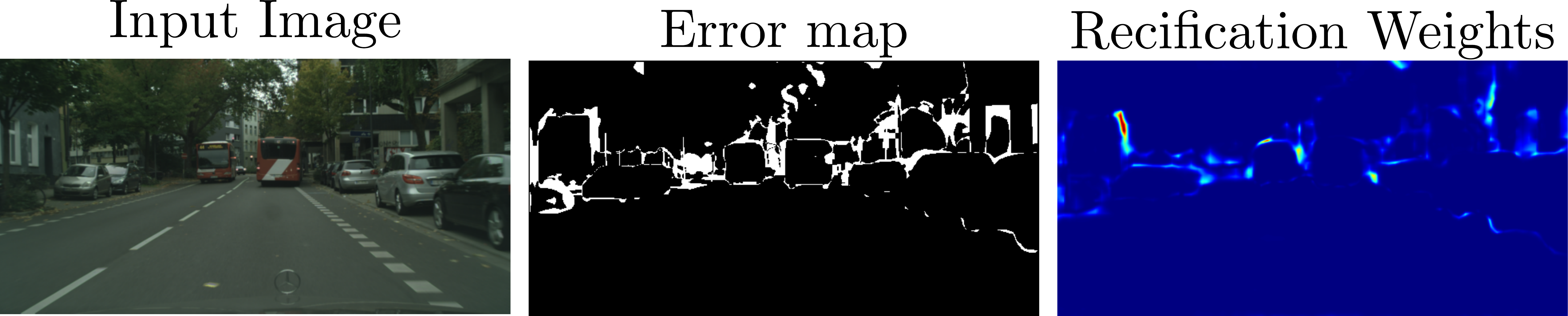}
    }
    \vspace{-3mm}
    \caption{(a) Proposed method for Multi-Target Domain Adaptation (MTDA). Feature stylization is performed to favor consistency across classifiers via pseudo-labelling. Classifier consistency is used to estimate pseudo-label quality and rectify the training loss. (b) We show the uncertainty map estimated from an input image and used for loss rectification (dark blue for high confidence). We observe that low confidence regions often correspond to errors.}%
    \label{fig:teaser}%
\end{figure}


Although a vast majority of UDA methods have been proposed for semantic segmentation in the single source and single target setting, in practical applications the assumption of a single target domain easily becomes vacuous. It is because the real world is more complex and target data can come from varying and different data distributions. For \eg, in autonomous driving applications, the vehicle might encounter cloudy, rainy, and sunny weather conditions in a span of a very short journey. In such cases, it would require to switch among various adapted models specialized for a certain weather condition. To prevent cumbersome deployment operations one can instead train and deploy a single model for all the target environments, which is otherwise known as \textit{Multi-Target Domain Adaptation} (MTDA). While in the context of object recognition MTDA has been explored in several works \cite{chen2019blending,gholami2020unsupervised,peng2019domain,roy2021curriculum,yang2020heterogeneous}, it is heavily understudied for semantic segmentation, with just a handful of existing works \cite{isobe2021multi, lee2022adas,saporta2021mtaf}. The prior works are either sub-optimal at fully addressing the target-target alignment \cite{saporta2021mtaf} or tackle it at a high computation overhead of explicit style-transfer \cite{isobe2021multi, lee2022adas}. We argue that explicit interactions between a pair of target domains are essential in MTDA for minimizing the domain gap across target domains. 

To this end, in this paper we present a novel MTDA framework for semantic segmentation that employs a \textit{self-training} strategy based on pseudo-labeling to induce better synergy between different domains.  
Self-training is a widely used technique consisting in comparing different predictions obtained from a single image to impose consistency in network's predictions. 
In our proposed method, illustrated in Fig.~\ref{fig:teaser} (a), we use an original image from one target domain (in yellow box) as the \textit{view} that generates the pseudo-label; while the second prediction is obtained with the very same target image but stylized with an image coming from a different target domain (in green box). Given this stylized feature, the network is then asked to predict the pseudo-label obtained from the original view. Unlike \cite{isobe2021multi} we use implicit stylization that does not need any externally trained style-transfer network, making our self-training end-to-end. Self-training not only helps the network to improve the quality of representations but also helps in \textit{implicit alignment} between target-target pairs due to cross-domain interactions. 

While our proposed self-training is well-suited for MTDA, it can still be susceptible to noisy pseudo-labels. To prevent the network from overfitting to noisy pseudo-labels when the domain-shift is large, we devise a \textit{cross-domain cooperative rectification} strategy that captures the disagreement in predictions from different classifiers. Specifically, our proposed method uses the predictions from multiple domain-specific classifiers to estimate the quality of pseudo-labels (see Fig.~\ref{fig:teaser} (b)), which are then weighted accordingly during self-training.  Thus, interactions between all the target domains are further leveraged with our proposed framework, which we call \textbf{Co}-oper\textbf{a}tive \textbf{S}elf-\textbf{T}raining (\emph{\ours}) for MTDA. 

\noindent\textbf{Contributions.} In summary, our contributions are three fold: \emph{(i)} We propose a \textit{self-training} approach for MTDA that synergistically combines pseudo-labeling and feature stylization to induce better cooperation between domains; \emph{(ii)} To reduce the impact of noisy pseudo-labels in self-training, we propose cross-domain \textit{cooperative objective rectification} that uses predictions from multiple domain-specific classifiers for better estimating the quality of pseudo-labels; and \emph{(iii)} We conduct experiments on several standard MTDA benchmarks and advance the state-of-the-art performance by non-trivial margins.

\vspace{-3mm}
\section{Related Works}
\label{sec:related}

\vspace{-3mm}
Our proposed method is most related to self-training and style-transfer, which we discuss in the following section.

\noindent \textbf{Self-training for Domain Adaptation}. Self-training in single-target domain adaptation (STDA) is a popular technique that involves generating pseudo-labels for the unlabeled target data and then iteratively training the model on the most confident labels. To that end, a plethora of UDA methods for semantic segmentation has been proposed \cite{kim2020learning,li2020content,li2019bidirectional,wang2020differential,zheng2020unsupervised,zheng2021rectifying,zou2018unsupervised} that use self-training due to its efficiency and simplicity. However, due to the characteristic error-prone nature of the pseudo-labeling strategy, the pseudo-labels cannot always be trusted and need a selection or correction mechanism. Most self-training methods differ in the manner in which the pseudo-labels are generated and selected. For instance, Zou \etal\tcite{zou2018unsupervised} proposed a class-balanced self-training strategy and used spatial priors, whereas in \tcite{zhang2021prototypical,zhang2019category} class-dependent centroids are employed to generate pseudo-labels. Most relevant to our approach are self-training methods\tcite{shen2019regularizing,zheng2020unsupervised,zheng2021rectifying} that rectify the pseudo-labels by measuring the uncertainty in predictions. Our proposed \emph{\ours} also derives inspirations from the STDA method \tcite{zheng2021rectifying}, but instead of ad-hoc auxiliary classifiers, we use different stylized versions of the same image and different target domain-specific classifiers, to compute the rectification weights. The majority of the STDA self-training methods do not trivially allow target-target interactions, which is very crucial for MTDA.

\noindent \textbf{Style-Transfer for Domain Adaptation}. Yet another popular technique in STDA that essentially relies on transferring \textit{style} (appearance) to make a source domain image look like a target image or vice versa. Assuming the semantic content in the image remains unchanged in the stylization process, and hence the pixel labels, \textit{target-like} source images can be used to train a model for the target domain. Thus, the main task becomes modeling the style and content in an image through an encoder-decoder-like network. In the context of STDA in semantic segmentation, Hoffman \etal\tcite{hoffman2018cycada} proposed \emph{CyCADA}, that incorporates cyclic reconstruction and semantic consistency to learn a classifier for the target data. Inspired by \emph{CyCADA} a multitude of STDA methods \cite{chang2019all,chen2019crdoco,li2018semantic,murez2018image,toldo2020unsupervised,wu2018dcan,yang2020fda,zhu2018penalizing} have been proposed which use style-transfer in conjunction with other techniques. Learning a good encoder-decoder style-transfer network introduces additional training overheads and the success is greatly limited by the reconstruction quality. Alternatively, style-transfer can be performed in the feature space of the encoder without explicitly generating the stylized image~\cite{tang2021crossnorm,zhou2021domain}. \emph{CrossNorm} \cite{tang2021crossnorm} explores this solution in the context of domain generalization to learn robust features. In \emph{\ours}, we adapt \emph{CrossNorm} to our self-training mechanism by transferring style across target domains to induce better synergy.

\noindent \textbf{Multi-target Domain Adaptation}. MTDA for semantic segmentation is an under-explored field with just a handful of existing works \cite{isobe2021multi, lee2022adas,saporta2021mtaf}. For instance, Saporta \etal\tcite{saporta2021mtaf} proposed an adversarial framework where source-target and target-target alignment is achieved through dedicated discriminators. They also introduced a multi-target knowledge transfer (\emph{MTKT}) approach where knowledge distillation (\emph{KD})\tcite{hinton2015distilling} is used to learn a domain-agnostic classifier from multiple domain-specific experts. On the other hand, the CCL\tcite{isobe2021multi} and ADAS~\cite{lee2022adas} rely on explicit style-transfer to tackle MTDA in semantic segmentation. Much like other style-transfer based STDA methods,~\cite{lee2022adas} uses an external network for explicitly transferring styles between domains. Instead, we rely on implicit style-transfer making our proposed \emph{\ours} easy to implement and end-to-end trainable. Additionally, we introduce a cooperative rectification technique which prevents over-fitting on imperfect pseudo-labels, making our method more robust. We empirically prove this effectiveness over\tcite{isobe2021multi,lee2022adas,saporta2021mtaf} through numerous experiments.

\vspace{-3mm}
\section{Methods}
\label{sec:methods}
\vspace{-2mm}
In this section we formally define the MTDA task and then we present the details of our proposed Cooperative Self-Training (\emph{\ours}) framework.
\vspace{-2mm}
\subsection{Preliminaries}
\label{sec:prelim}
\vspace{-2mm}

\begin{figure*}[t]
\centering
\includegraphics[width=0.99\textwidth]{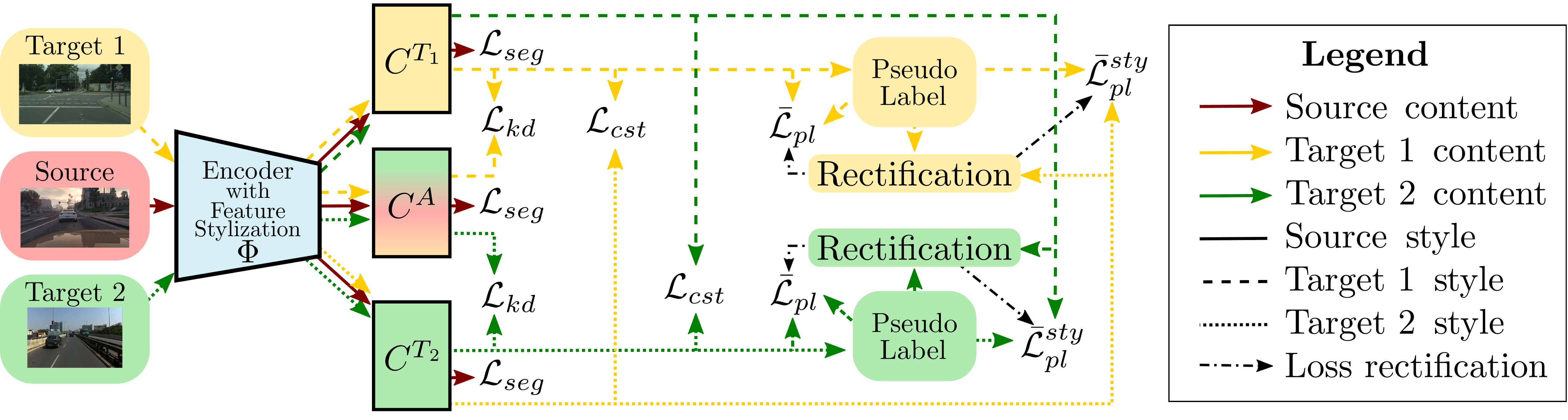}
\vspace{-0.3cm}\caption{Illustration  of the proposed \emph{\ours} approach in the case of two target domains. Domain-specific classifiers are distilled to learn a domain-agnostic classifier. Style-transfer is used in the encoder network to induce cooperation between the different classifiers and rectify the pseudo-labeling losses.}
\label{fig:warmup}
\end{figure*} 

\noindent \textbf{Problem Definition and Notations}. In the multi-target domain adaptation (MTDA) task, we assume that we have at our disposal $N\tsource$ labeled instances from a source domain data set $\data\tsource = \{(\vx\tsource_n, \vy\tsource_n)\}^{N\tsource}_{n=1}$ where $\vx\tsource \in \mathbb{R}^{H \times W \times 3}$ are input images with their corresponding one-hot ground truth labels $\vy\tsource \in \mathbb{R}^{H \times W \times K}$, assigned to each pixel in the $H \times W$ spatial grid belonging to one of the $K$ semantic classes. Moreover, there are a total of $M$ unlabeled target domains $\{\target_1, \dots, \target_M\}$ where each target domain $\target_i$ comprises of an unlabeled data set $\data\ttarget{i} = \{\vx\ttarget{i}_n\}^{N\ttarget{i}}_{n=1}$, with $\vx\ttarget{i} \in \mathbb{R}^{H \times W \times 3}$ representing the target images and $N\ttarget{i}$ being the number of unlabeled instances. Following standard MTDA protocols, we assume that the marginal distributions between every pair of available domains differ, under the constraint of underlying semantic concept remaining the same. The goal of MTDA is to learn a single network $f = C \circ \Phi$ using $\{\bigcup^{M}_{i=1}\data\ttarget{i}\} \cup \data\tsource$ that can segment samples from any target domain, where $C$ and $\phi$ are the classifier and the backbone encoder networks, respectively. While we consider that the domain information is known at training time, the domains labels of the images during inference are unknown.

\noindent \textbf{Overall Framework}. To address the MTDA, we operate in two stages. In the first stage we aim to learn target domain-specific classifiers with \textit{adversarial adaptation} \cite{tsai2018learning,vu2018advent} that aligns features between a given source-target domain pair.
The first stage results in the network parameters that enable even better alignment in the subsequent stage. In this second stage, we adopt a pseudo-label based \textit{cooperative self-training} strategy to further align the target domains. In particular, our proposed self-training strategy enforces consistency among the target domain-specific classifiers, allowing maximal interaction among the different target domains. Importantly, our cooperative training also incorporates a \textit{threshold-free rectification} term that prevents overfitting to noisy pseudo-labels. Finally, we use \textit{knowledge distillation} to distill all the learned information from domain-specific classifiers to a domain-agnostic classifier that can be used to segment a test image from any target domain, thereby alleviating the need for domain-id during inference.

\noindent\textbf{Adversarial Warm-up}. This marks the first stage, where we follow \cite{saporta2021mtaf} for initializing our framework in order to obtain an encoder network $\Phi$ that is shared among all the target domains, and $M$ distinct target domain-specific classifiers $\{C\ttarget{i}| \forall i\in\{1, \dots, M\}\}$. Concurrently, we also initialize $M$ target domain-specific discriminators $\{D\ttarget{i}| \forall i\in\{1, \dots, M\}\}$ to learn a classifier that is invariant for a specific source-target pair. To recap, in adversarial warm-up stage the discriminator $D\ttarget{i}$ is trained to distinguish between the source and target $\target_i$ predictions whereas the network $f\ttarget{i} = C\ttarget{i} \circ \Phi$ is trained to fool the $D\ttarget{i}$. Note that unlike the original work in \cite{ganin2015unsupervised}, the output from the classifier is given as an input to the domain discriminator \cite{saporta2021mtaf,tsai2018learning}. Additionally, for the source samples we employ the standard supervised cross-entropy loss, which is used to train every $f\ttarget{i}$. Overall, for a given source-target pair $(\source, \target_i)$ the discriminator $D\ttarget{i}$ is trained with the objective: 
\begin{equation}
\begin{split}
\label{eqn:warm-up-dis}
    \mathcal{L}_{D\ttarget{i}} = \mathcal{L}_\mathrm{bce} \big(D\ttarget{i}(C\ttarget{i}(\Phi(\vx\tsource))), 1\big) + \\  \mathcal{L}_\mathrm{bce}\big(D\ttarget{i}(C\ttarget{i}(\Phi(\vx\ttarget{i}))), 0\big)
\end{split}
\end{equation}
where $\mathcal{L}_\mathrm{bce}$ stands for the \textit{binary cross-entropy} loss. Simultaneously, the network $f\ttarget{i}$ is trained along with the source segmentation loss and adversarial loss as:
\begin{equation}
\begin{split}
\label{eqn:warm-up-feat}
    \mathcal{L}_{f\ttarget{i}} = \mathcal{L}_\mathrm{ce} \big(C\ttarget{i} (\Phi(\vx\tsource)), \vy\tsource\big) + \\  \lambda_{adv} \mathcal{L}_\mathrm{bce} \big(D\ttarget{i}(C\ttarget{i}(\Phi(\vx\ttarget{i}))), 1\big)
\end{split}
\end{equation}
where $\mathcal{L}_\mathrm{ce}$ is the supervised cross-entropy loss for the source data and $\lambda_{adv}$ is a hyperparameter to balance the losses. In the adversarial warm-up stage we alternatively minimize $\mathcal{L}_\mathrm{D\ttarget{i}}$ and $\mathcal{L}_\mathrm{f\ttarget{i}}$ for every source-target pairs.

\vspace{-3mm}
\subsection{Cooperative Self-Training (\emph{\ours})}
\label{sec:coll-st}
\vspace{-3mm}
The goal of the second stage is to refine the image representation learned in the adversarial warm-up stage. We devise a self-training approach with the usage of pseudo-labels that iteratively improves the predictions of the model on the unlabeled data. 






\noindent \textbf{Pseudo-labelling.} In our framework for the MTDA, we have $m$ specialized target domain-specific classifiers, with each classifier $C\ttarget{i}$ trained to handle data coming from the corresponding domain $\target_i$. We exploit these specialized classifiers to generate pseudo-labels (PLs) for the target samples in their respective target domains. Specifically, given the $n$\ths image $\vx\ttarget{i}_n$ from the target domain $i$, we use the network $f\ttarget{i}$ to predict the segmentation map  $[\hat{\mpred}\ttarget{i}_n(\textbf{k})]_{\textbf{k} \in [H] \times [W] \times [K]} = C\ttarget{i}(\Phi(\vx\ttarget{i}_n))$ and compute the pseudo-label as:
\begin{equation}
\label{eqn:one-hot}
    \hat{\vy}\ttarget{i}_n = \mbf{e}_k \big(\argmax_{k} [\hat{\mpred}\ttarget{i}_n(\textbf{k})]_{\textbf{k} \in [H] \times [W] \times [K]}\big), 
\end{equation}
where $\mbf{e}_k(.)$ denotes the one-hot encoding operator and $\hat{\vy}\ttarget{i}_n \in \mathbb{R}^{H \times W \times K}$. The PL is computed at the beginning of the second stage and is updated every $n_b$ iterations. This PL is then used to self-supervise the corresponding $f\ttarget{i}$ network with a \textit{cross entropy} loss:
\begin{equation}
\label{eqn:pl}
\mathcal{L}_\mathrm{pl} = \mathcal{L}_\mathrm{ce} (\hat{\mpred}\ttarget{i}_n, \hat{\vy}\ttarget{i}_n),
\end{equation}

However, this formulation suffers from two main issues. First, the PLs act only on the same domain-specific classifier $C\ttarget{i}$ corresponding to the domain of input images. Hence, it does not induce any synergy between the different classifiers. Second, since the PLs can be noisy, using the pseudo-labeling objective in Eq.~(\ref{eqn:pl}) can lead to detrimental behaviour.
To address these two issues and further benefit from our PLs, we introduce a self-training technique that is realized by leveraging feature stylization~\cite{tang2021crossnorm}.

\noindent\textbf{Style-Transfer for Cooperative Self-Training}. To benefit from the self-training objective in Eq.~(\ref{eqn:pl}), one requires to obtain the predictions from a view $t(\vx\ttarget{i}_n)$ and enforce its predictions to match with that of $\hat{\vy}\ttarget{i}_n$, where $t(.)$ is any stochastic transformation. Indeed, such a consistency-based training strategy has successfully been applied in the semi-supervised learning literature \cite{sohn2020fixmatch}. However, finding optimal transformations is not trivial and varies between data sets and even tasks. In this work, we resort to a data-driven transformation policy that is based on style-transfer~\cite{ulyanov2016instance}. Style-transfer consists in transferring the``style'' (appearance) from one image to another. Concretely, in our case, the transformation $t(.)$ is a style-transfer operation that essentially applies the style of an image $\vx\ttarget{j}$ to the image $\vx\ttarget{i}$, where $i \neq j$. The style transformed image $\vx\ttarget{i \to j}$ can in essence be regarded as a \textit{virtual} image that \textit{appears} to come from $\target_j$ but having the content structure of $\target_i$. Therefore, for the $n$\ths sample $\vx\ttarget{i \to j}$ we obtain the prediction from $f\ttarget{j}$ and optimize it to be close to $\hat{\vy}\ttarget{i}_n$. In this way our PL from a given target domain-specific classifier can be used to supervise another domain-specific classifier, enforcing better consistency between pairs of target domains. Moreover, we thereafter show how style-transfer is instrumental in rectifying the objective in Eq.~(\ref{eqn:pl}) according to an estimated confidence score. We now describe how we use style-transfer to improve self-training in the MTDA setting.

Style-transfer in the pixel space, with separately trained encoder-decoder network, has very recently been used for the MTDA work~\cite{lee2022adas}. To avoid such costly, and often sub-optimal, image generation with the pixel-space style-transfer methods, we perform style-transfer in the intermediate feature space of the encoder network. In particular, we adapt cross normalization (\emph{CrossNorm})~\cite{tang2021crossnorm} in our MTDA setting and use it as a means of exchanging feature statistics, and hence style, across different domains. More precisely, our Cross-Domain Normalization (\emph{CrossDoNorm}) performs style-transfer by exchanging \textit{style vectors} between two target domain images, which are computed from the channel-wise mean and standard deviation of the features maps. Exchange of style vectors is deemed sufficient for style-transfer by prior works~\cite{ulyanov2016instance} who show that these statistics encode the image style and that style-transfer can be obtained through a simple re-normalization.

Given a pair of images ($\vx\ttarget{i}, \vx\ttarget{j}$) from the target domains $\target_{i}$ and $\target_{j}$, we extract their corresponding features from the $l$\ths layer of the encoder as $\vz\ttarget{i}_l = \Phi_l(\vx\ttarget{i})$ and $\vz\ttarget{j}_l = \Phi_l(\vx\ttarget{j})$. From these intermediate feature maps, we compute the corresponding channel-wise means ($\bm{\mu}\ttarget{i}_l, \bm{\mu}\ttarget{j}_l$) and standard deviations ($\bm{\sigma}\ttarget{i}_l, \bm{\sigma}\ttarget{j}_l$), such that $\bm{\mu}_l \in \mathbb{R}^k$ and $\bm{\sigma}_l \in \mathbb{R}^k$ with $k$ being the number of channels in the layer $l$. For instance, \emph{CrossDoNorm} first standardizes the features with its own channel-wise statistics, \eg, ($\bm{\mu}\ttarget{i}_l, \bm{\sigma}\ttarget{i}_l$) for $\vz\ttarget{i}_l$, and then re-normalizes with the statistics from the other domain ($\bm{\mu}\ttarget{j}_l, \bm{\sigma}\ttarget{j}_l$) to obtain stylized features $\vz\ttarget{i \to j}_l$. The \emph{CrossDoNorm} can be done symmetrically resulting into stylized features that are computed as:
\begin{equation}
\begin{split}
\label{eqn:crossdonorm}
\vz\ttarget{i \to j}_l=  \bm{\sigma}\ttarget{j}_{l} \frac{\vz\ttarget{i}_{l} - \bm\mu\ttarget{i}_{l}}{\bm\sigma\ttarget{i}_{l}} + \bm\mu\ttarget{j}_{l} \\ 
\vz\ttarget{j \to i}_l=  \bm{\sigma}\ttarget{i}_{l} \frac{\vz\ttarget{j}_{l} - \bm\mu\ttarget{j}_{l}}{\bm\sigma\ttarget{j}_{l}} + \bm\mu\ttarget{i}_{l}
\end{split}
\end{equation}

Our \emph{CrossDoNorm} can ideally perform feature stylization at multiple layers in the encoder network. Next, the above computed stylized feature $\vz\ttarget{i \to j}_l$ is then given as input to the subsequent layers of the network, with the final prediction map $\hat{\mpred}\ttarget{i \to j}$ obtained from the classifier $C\ttarget{j}$. With the PL $\hat{\vy}\ttarget{i}$ generated from $\vz\ttarget{i}_l$ by the original domain classifier $C\ttarget{i}$, the other classifier $C\ttarget{j}$, along with the encoder $\Phi$, is then trained in a supervised manner:
\begin{equation}
\begin{split}
\label{eqn:style-pl}
  \mathcal{L}_\mathrm{pl}^\mathrm{sty}(\vx\ttarget{i}, \vx\ttarget{j}) = \mathcal{L}_\mathrm{ce}(\hat{\mpred}\ttarget{i \to j}, \hat{\vy}\ttarget{i}) \\ 
  \mathcal{L}_\mathrm{pl}^\mathrm{sty}(\vx\ttarget{j}, \vx\ttarget{i}) = \mathcal{L}_\mathrm{ce}(\hat{\mpred}\ttarget{j \to i}, \hat{\vy}\ttarget{j})
\end{split}
\end{equation}

Since it has been shown in the literature that training with soft-labels improves the learning ability of the network \cite{hinton2015distilling}, we use a soft-version of the $\mathcal{L}^\mathrm{sty}_\mathrm{pl}$ loss described in Eq.~(\ref{eqn:style-pl}). In other words, we further enforce consistency in predictions between two domain-specific classifiers by optimizing the KL-divergence objective between the cross-domain stylized prediction $\hat{\mpred}\ttarget{i \to j}$ and the original target domain prediction $\hat{\mpred}\ttarget{i}$ instead of PLs as:
\begin{equation}
\label{eqn:cst}
  \mathcal{L}_\mathrm{cst}(\vx\ttarget{i}, \vx\ttarget{j}) = \mathcal{L}_\mathrm{kl}(\hat{\mpred}\ttarget{i \to j}, \hat{\mpred}\ttarget{i})
\end{equation}

Additionally, our \emph{CrossDoNorm} also acts as an implicit data augmentation method in the feature space. As the style information is mainly manifested in the low level features of the encoder, to prevent over-regularization we only apply the \emph{CrossDoNorm} in the initial layers of the encoder.

\noindent\textbf{Cooperative Objective Rectification.} The PLs generated during the refinement process can be very noisy due to domain-shift, leading to degradation of representations. To tackle this shortcoming of self-training, we propose our cooperative objective rectification method that takes into account the uncertainty in the model predictions. This uncertainty in predictions for a given sample $\vx\ttarget{i}_n$ is measured by combining the predictions obtained from all the target domain classifiers. More precisely, considering $\vx\ttarget{i}_n$, we compute the consistency scores between the prediction from the $C\ttarget{i}$ and the predictions from the other domain-specific classifiers on the stylized features of $\vx\ttarget{i}_n$. Following~\cite{zheng2021rectifying}, we use the KL-divergence between a pair of predictions as a measure of consistency. 
Lower the consistency, less reliable is the corresponding PL. Finally, this consistency score is then used as a weight to re-weight the self-training loss introduced in the Eq.~(\ref{eqn:pl}). The rectified self-training loss corresponding to Eq.~(\ref{eqn:pl}) is given as:
\begin{equation}
\bar{\mathcal{L}}_\mathrm{pl}(\vx\ttarget{i}) = w_i\mathcal{L}_\mathrm{ce}(\hat{\mpred}\ttarget{i}, \hat{\vy}\ttarget{i})
\end{equation}
where the weight value  $w_i$ is the averaged consistency scores obtained with the predictions between $C\ttarget{i}$ and the rest of the classifiers \{$C\ttarget{1}, \dots, C\ttarget{M}$\}$\setminus C\ttarget{i}$ as:
\begin{equation}
\label{eqn:rectification-weight}
  w_i = \frac{1}{M-1}  \sum_{j=1, j\neq i}^M \exp{(-\mathcal{L}_\mathrm{kl}(\hat{\mpred}\ttarget{i}, \hat{\mpred}\ttarget{i \to j})})
\end{equation}
where the exponential function exp($\cdot$) is used here to map the KL divergence that range in $[0,+\infty[$ to weights values in ]0,1]. Contrary to \cite{zheng2021rectifying} our uncertainty score is obtained by considering the predictions from all classifier pairs, against just using a single pair of classifiers. Also, different from many pseudo-labeling approaches described in Sec.~\ref{sec:related}, our re-weighting formulation is not based on thresholding and therefore avoids manual hyperparameter tuning.
Similarly, the cross-domain pseudo labeling losses introduced in Eq.~(\ref{eqn:style-pl}) are rectified as: 
\begin{equation}
\begin{split}
\label{eqn:pl_cross}
\bar{\mathcal{L}}^\mathrm{sty}_\mathrm{pl}(\vx\ttarget{i}, \vx\ttarget{j}) = w_i\mathcal{L}_\mathrm{ce}(\hat{\mpred}\ttarget{i \to j}, \hat{\vy}\ttarget{i}) \\ 
\bar{\mathcal{L}}^\mathrm{sty}_\mathrm{pl}(\vx\ttarget{j}, \vx\ttarget{i}) = w_j\mathcal{L}_\mathrm{ce}(\hat{\mpred}\ttarget{j \to i}, \hat{\vy}\ttarget{j})
\end{split}
\end{equation}

\noindent\textbf{Knowledge Distillation}. As our end goal is to be able to predict test samples coming from any target domain, we also learn an additional domain-agnostic classifier $C^\mathrm{A}$. We use the source samples to train $C^\mathrm{A}$ in addition to the supervised segmentation objective given in Eq.~(\ref{eqn:warm-up-feat}) as:
\begin{equation}
\begin{split}
\label{eqn:seqall}
  \mathcal{L}_\mathrm{seg} = \sum_{i=1}^{M} \mathcal{L}_\mathrm{ce}(C\ttarget{i}(\Phi(\vx\tsource)), \vy\tsource)\\ +\mathcal{L}_\mathrm{ce}(C^\mathrm{A}(\Phi(\vx\tsource)), \vy\tsource)
 \end{split}
\end{equation}

In order to distill the information learned by the domain-specific classifiers $C\ttarget{i}$ into the domain-agnostic classifier $C^\mathrm{A}$ we use knowledge distillation (KD) as in \cite{isobe2021multi,saporta2021mtaf}. For every target domain sample, we enforce consistency between the prediction from the corresponding domain-specific classifier and the domain-agnostic one using the KL divergence. The KD loss for a given $\target_i$ domain is given as: 
\begin{equation}
\label{eqn:kd}
\mathcal{L}_\mathrm{kd} = \mathcal{L}_\mathrm{kl}(C^{\mathrm{A}}(\Phi(\vx\ttarget{i})),\hat{\vy}\ttarget{i})
\end{equation}
where only the weights of the $C^\mathrm{A}$ is only updated during the optimization of Eqn.~\ref{eqn:kd}. We use the domain-agnostic classifier $C^\mathrm{A}$ during inference.

\noindent\textbf{Overall Training.} The final objective to train our proposed \emph{\ours} is given by summing all the unary and pairwise losses previously described: 
\begin{equation}
\begin{split}
\label{eqn:final-objective}
  &\mathcal{L}_\mathrm{\emph{\ours}}=\sum_{(\vx\tsource, \vy\tsource) \in \data\tsource}  \mathcal{L}_\mathrm{seg}(\vx\tsource, \vy\tsource) + \\
  &\sum_{i=1}^M \sum_{\vx\ttarget{i}\in \data\ttarget{i}} \Big[ \frac{1}{M}\mathcal{L}_\mathrm{kd}(\vx\ttarget{i}) +\bar{\mathcal{L}}_\mathrm{pl}(\vx\ttarget{i}) + \\
  &\frac{1}{M-1}\sum_{\substack{j=1\\j\neq i}}^M\sum_{\vx\ttarget{j}\in \data\ttarget{j}}
  \bar{\mathcal{L}}_\mathrm{pl}^\mathrm{sty}(\vx\ttarget{i},\vx\ttarget{j})+  \mathcal{L}_\mathrm{cst}(\vx\ttarget{i},\vx\ttarget{j})\Big] 
\end{split}
\end{equation}
Note that, the KD loss and pairwise losses are normalized by $M$ and $M-1$ to preserve the source-target balance when varying the number of target domains.
\vspace{-5mm}



\section{Experiments}

\vspace{-3mm}
\subsection{Experimental set-up}
\vspace{-3mm}
\noindent\textbf{Datasets.}
We conduct experiments on two standard benchmarks for MTDA in semantic segmentation. These two benchmarks have been derived from four semantic segmentation datasets, namely the synthetic \emph{GTA5}~\cite{richter2016playing} and the real world \emph{Cityscapes}~\cite{cordts2016cityscapes}, \emph{Mapillary}~\cite{neuhold2017mapillary} and \emph{IDD}~\cite{varma2019idd}. Note that the datasets are varying in size as in the Mapillary is six times bigger than the Cityscapes, and thrice as big as the IDD. More details can be found in the supplement.

\noindent\textbf{Benchmarks.}
The benchmarks for MTDA in semantic segmentation differ in the way the class labels are mapped across the datasets. They are: (i) the \emph{7-classes} benchmark, introduced in \cite{saporta2021mtaf}, which considers 7 classes and down-samples the images to a resolution of $640 \times 320$ both for training and evaluation; and (ii) the \emph{19-classes} benchmark, introduced in \cite{isobe2021multi}, which operates at higher resolution of $1024 \times 512$. Both benchmarks use several combinations of the four datasets to create four \emph{Synthetic to Real} scenarios 
and one \emph{Real to Real} scenario. 

\noindent\textbf{Metrics.} We report the standard intersection over union (IoU) for every class and the mean-IoU (mIoU) for each target domain. Whereas, to obtain a single overall score in the MTDA, we average the mIoU across all the target domains.

\noindent\textbf{Baselines.}
In our experiments, we compare with the state-of-the-art methods: Multi-Target Knowledge Transfer (\emph{MTKT}) \cite{saporta2021mtaf}, Collaborative Consistency Learning (\emph{CCL})~\cite{isobe2021multi} and A Direct Adaptation Strategy (\emph{ADAS})~\cite{lee2022adas}. We compare with these methods on the settings adopted in the corresponding papers: \emph{7-classes} for \emph{MTKT} and \emph{19-classes} for \emph{CCL}. 
We also include an approach, introduced in \cite{saporta2021mtaf} and referred to as \emph{Multi-Discriminator}, where a single classifier is trained using multiple domain-specific discriminator.
In addition, we follow \cite{isobe2021multi,saporta2021mtaf} and include two baselines based on a single-target domain adaptation method. In \emph{Individual}, an adversarial approach  \cite{vu2018advent} is trained separately on every target dataset. At inference time, the target images are tested by the corresponding domain-specific model. In \emph{Data combination}, we treat the union of all the target domains as a single target domain. For these two baselines, we report the results provided in \cite{isobe2021multi,saporta2021mtaf}. 

\noindent\textbf{Implementation Details.}
To be fairly comparable, we adopt the very same network architecture as in the baseline~\cite{saporta2021mtaf}, except we use the modified version of ResNet101 based DeepLab-V2~\cite{chen2017deeplab} that contains dropout layers~\cite{zhang2021prototypical,zheng2021rectifying}. Due to lack of space we report the rest of the implementation details in the supplement. 
\setlength{\tabcolsep}{4pt}
\begin{table*}[t]
\def\arraystretch{.85}
\begin{center}

\resizebox{\textwidth}{!}{
    \begin{tabular}{c|c|c c c c c c c |c|c}
    \hline
    \noalign{\smallskip}
    \multicolumn{11}{c}{\bf GTA5 $\rightarrow$ Cityscapes + IDD}\\ 
    \noalign{\smallskip}
    \hline
    \bf Method & \bf Target & \bf flat & \bf constr & \bf object & \bf nature & \bf sky & \bf human & \bf vehicle & \bf mIoU & \bf Avg. \\ \hline
    
    
    
    \multirow{2}*{\emph{Individual}~\cite{vu2018advent}} & C & 93.5 & 80.5 & 26.0 & 78.5 & 78.5 & 55.1 & 76.4 & 69.8 & \multirow{2}*{67.5} \\  
    ~ & I & 91.2 & 53.1 & 16.0 & 78.2 & 90.7 & 47.9 & 78.9 & 65.1 & ~ \\ \hline
    
    \multirow{2}*{\emph{Data Comb.}~\cite{vu2018advent}} & C  & 93.9 & 80.2 & 26.2 & 79.0 & 80.5 & 52.5 & 78.0 & 70.0 & \multirow{2}*{67.4} \\  
    ~ & I  & 91.8 & 54.5 & 14.4 & 76.8 & 90.3 & 47.5 & 78.3 & 64.8 & ~ \\ \hline

    \multirow{2}*{\emph{Multi-Dis}~\cite{saporta2021mtaf}} & C  & 94.3 & 80.7 & 20.9 & 79.3 & 82.6 & 48.5 & 76.2 & 68.9 & \multirow{2}*{67.3} \\  
    ~ & I & 92.3 & 55.0 & 12.2 & 77.7 & 92.4 & 51.0 & 80.2 & 65.7 & ~ \\ \hline
    
    \multirow{2}*{\emph{MTKT}~\cite{saporta2021mtaf}} & C  & 94.5 & 82.0 & 23.7 & 80.1 & 84.0 & 51.0 & 77.6 & 70.4 & \multirow{2}*{68.2} \\  
    ~ & I  & 91.4 & 56.6 & 13.2 & 77.3 & 91.4 & 51.4 & 79.9 & 65.9 & ~ \\ \hline
    
    \multirow{2}*{\emph{ADAS}~\cite{lee2022adas}($1024\times512$)} & C  & \textbf{95.1} & 82.6 & \textbf{39.8} & \textbf{84.6} & 81.2 & \textbf{63.6} & \textbf{80.7} & \textbf{75.4} & \multirow{2}*{71.2} \\  
    ~ & I  & 90.5 & \textbf{63.0} & \textbf{22.2} & 73.7 & 87.9 & 54.3 & 76.9 & 66.9 & ~ \\ \hline
    
    \multirow{2}*{\emph{\ours} (Ours)} & C  & 94.7 & \textbf{82.9} & 25.4 & 82.2 & \textbf{88.2} & 54.4 & 80.5 & 72.6 & \multirow{2}*{\textbf{71.3}} \\  
    ~ & I & \textbf{94.2} & 61.5 & 20.0 & \textbf{82.7} & \textbf{93.4} & \textbf{55.5} & \textbf{82.6} & \textbf{70.0} & ~ \\ \hline
    
    \end{tabular}
}
\end{center}
\vspace{-0.4cm}
\caption{Comparison with State-of-the-art on the \emph{7-classes} benchmark using the GTA5 $\rightarrow$ Cityscapes + IDD  configuration. }
\label{table:G2CI}
\end{table*}
\setlength{\tabcolsep}{1.4pt}

\setlength{\tabcolsep}{4pt}
\begin{table*}[t]
\def\arraystretch{.9}
\begin{center}

\resizebox{\textwidth}{!}{
    \begin{tabular}{c|c|c c c c c c c c c c c c c c c c c c c|c|c}
    \hline
    \noalign{\smallskip}
    \multicolumn{23}{c}{\bf GTA5 $\rightarrow$ Cityscapes + IDD}\\
    \noalign{\smallskip}
    \hline
    \bf Method & \rotatebox{90}{\bf Target} & \rotatebox{90}{\bf road} & \rotatebox{90}{\bf  sidewalk} & \rotatebox{90}{\bf  building} & \rotatebox{90}{\bf walk} & \rotatebox{90}{\bf  fence} & \rotatebox{90}{\bf  pole} & \rotatebox{90}{\bf light} & \rotatebox{90}{\bf sign} & \rotatebox{90}{\bf veg} & \rotatebox{90}{\bf terrain} & \rotatebox{90}{\bf sky} & \rotatebox{90}{\bf person} & \rotatebox{90}{\bf rider} & \rotatebox{90}{\bf car} & \rotatebox{90}{\bf truck} & \rotatebox{90}{\bf bus} & \rotatebox{90}{\bf train} & \rotatebox{90}{\bf motor} & \rotatebox{90}{\bf bike} & \bf mIoU & \bf Avg.\\ \hline

    \multirow{2}*{\emph{Individual}~\cite{vu2018advent}} & C & 88.8 & 23.8 & 81.5 & 27.7 & 27.3 & 31.7 & 33.2 & 22.9 & 83.1 & 27.0 & 76.4 & 58.5 & \textbf{28.9} & 84.3 & 30.0 & 36.8 & 0.3 & 27.7 & \textbf{33.1} & 43.3 & \multirow{2}*{43.5} \\  
    ~ & I & 94.1 & 24.4 & \textbf{66.1} & 31.3 & 22.0 & 25.4 & 9.3 & 26.7 & 80.0 & \textbf{31.4} & \textbf{93.5} & 48.7 & 43.8 & 71.4 & 49.4 & 28.5 & \textbf{0} & 48.7 & \textbf{34.3} & 43.6 & ~ \\ \hline

    \multirow{2}*{\emph{Data Comb.}\cite{vu2018advent}} & C & 86.1 & 32.0 & 79.8 & 24.3 & 22.3 & 28.5 & 27.9 & 14.3 & 85.1 & 29.8 & 79.9 & 56.1 & 20.5 & 77.7 & \textbf{34.4} & 35.2 & \textbf{0.7} & 18.2 & 13.1 & 40.3 & \multirow{2}*{41.2}  \\  
    ~ & I & 92.8 & 23.4 & 60.9 & 25.8 & 23.4 & 24.1 & 8.6 & 32.2 & 77.5 & 26.8 & 92.3 & 48.0 & 41.0 & 74.4 & 48.4 & 17.7 & \textbf{0} & 52.5 & 28.2 & 42.0 & ~ \\ \hline

    \multirow{2}*{\emph{CCL}~\cite{isobe2021multi}} & C & \textbf{90.3} & 34.0 & \textbf{82.5} & 26.2 & 26.6 & 33.6 & 35.4 & 21.5 & 84.7 & 39.8 & 81.1 & 58.4 & 25.8 & \textbf{84.5} & 31.4 & 45.4 & 0 & 29.9 & 24.7 & 45.0  & \multirow{2}*{45.5} \\  
    ~ & I & \textbf{95.0} & 30.5 & 65.6 & 29.4 & 23.4 & 29.2 & 12.0 & 37.8 & 77.3 & 31.3 & 91.9 & \textbf{52.4} & 48.3 & 74.9 & 50.1 & 36.6 & \textbf{0} & 56.1 & 32.4 & 46.0 & ~ \\ \hline
    
    \multirow{2}*{\emph{ADAS}~\cite{lee2022adas}} & C & - & - & - & - & - & - & - & - & - & - & - & - & - & - & - & - & - & - & - & 45.8  & \multirow{2}*{46.1} \\  
    ~ & I & - & - & - & - & - & - & - & - & - & - & - & - & - & - & - & - & - & - & - & 46.3  & ~ \\ \hline
    
    \multirow{2}*{\emph{\ours} (Ours)} & C & 81.7 & \textbf{38.3} & 71.0 & \textbf{33.3} & \textbf{30.7} & \textbf{35.1} & \textbf{38.2} & \textbf{37.6} & \textbf{86.4} & \textbf{46.9} & \textbf{81.9} & \textbf{63.4} & 27.4 & \textbf{84.5} & 29.4 & \textbf{45.6} & 0.3 & \textbf{32.6} & 31.3 & \textbf{47.1}  & \multirow{2}*{ \bf 48.2} \\  
    ~ & I & 85.7 & \textbf{36.1} & 65.1 & \textbf{33.2} & \textbf{23.7} & \textbf{32.8} & \textbf{19.0} & \textbf{62.9} & \textbf{82.5} & 29.5 & 91.8 & 52.1 & \textbf{55.3} & \textbf{83.4} & \textbf{62.9} & \textbf{46.1} & \textbf{0} & \textbf{55.5} & 18.5 & \textbf{49.3} & ~\\ \hline
    
    \end{tabular}
}
\end{center}\vspace{-0.4cm}
\caption{Comparison with State-of-the-art on the \emph{19-classes} benchmark using the GTA5 $\rightarrow$ Cityscapes + IDD  configuration.}
\label{table:G2CI19}
\vspace{-3mm}
\end{table*}
\setlength{\tabcolsep}{1.4pt}

\subsection{Comparison with State-of-the-art: Syn to Real}

\noindent\textbf{Quantitative Comparison.}
We provide a detailed comparison with state-of-the-art on the \emph{7-classes} benchmark using the GTA5 to Cityscapes and IDD setting. Results are reported in Tab.~\ref{table:G2CI}. Overall, we can observe that our method outperforms all the other baselines.  In terms of average mIoU, \emph{\ours} outperforms \emph{MTKT} with a $3.1\%$ margin. This gain is remarkable considering that \emph{MTKT} improves the \emph{Individual} baseline trained on Cityscapes by only $0.7\%$. Besides, \emph{\ours} outperforms \emph{ADAS} by $0.1\%$ even though \emph{ADAS} use a higer image resolution than \emph{\ours} and the other baselines. We can observe gains with respect to \emph{MTKT} in both small objects such as \emph{human} ($54.4\%$ vs $51.0\%$ on Cityscapes) and background classes such as sky ($88.2\%$ vs $84.0\%$ on Cityscapes). One noticeable point is that the IDD dataset seems more challenging since \emph{Individual} obtains lower performance on this dataset. 
Similarly, the \emph{Multi-Dis}, \emph{MTKT} and \emph{ADAS} obtain mIoUs of $65.7\%$, $65.9\%$ and $66.9\%$ respectively which are much lower than on Cityscapes ($68.9\%$, $70.4\%$ and $75.4\%$). However, with \emph{\ours}, which uses consistency training and cooperative objective rectification, we improve \emph{MTKT} and \emph{ADAS} performance by $4.1\%$ and $3.1\%$ obtaining a mIoU score of $70.0\%$.


We now provide experimental results on the \emph{19-classes} benchmark using the same GTA5 to Cityscapes and IDD setting. Results are reported in Tab.~\ref{table:G2CI19}. First, we observe that all methods have lower scores compared to the \emph{7-classes} since the high number of classes makes the task more difficult. Nevertheless, we observe that \emph{\ours} outperforms all the other approaches on almost all the classes and domains. Compared to \emph{CCL} and \emph{ADAS}, we observe that \emph{\ours} obtains better average mIoU ($+2.7\%$ and $+2.1\%$ respectively) and that the gain is mostly explained by better performances on difficult classes such as \emph{fence} or \emph{sign} and \emph{bus} that largely compensate the drop on the road class. 

\begin{figure*}[h]%
    \centering
    \resizebox{\textwidth}{!}{
    \begin{tabular}{cccc cccc}
    \multicolumn{4}{c}{\bf Cityscapes}  & \multicolumn{4}{c}{\bf IDD}\\
    Input& GT & \emph{MTKT} & \emph{\ours} (Ours) & Input& GT & \emph{MTKT} & \emph{\ours} (Ours)\\
       \includegraphics[height=1.0cm]{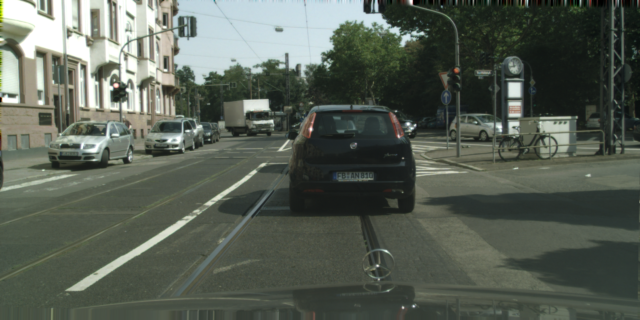} & \includegraphics[height=1.0cm]{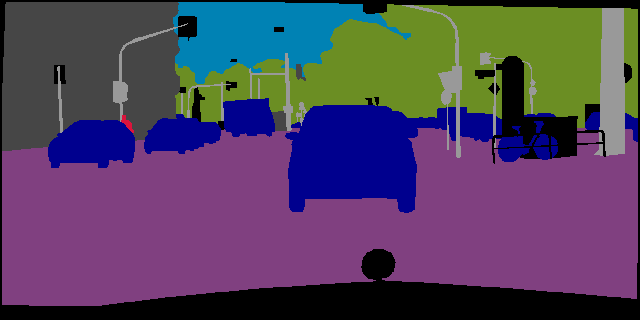} & \includegraphics[height=1.0cm]{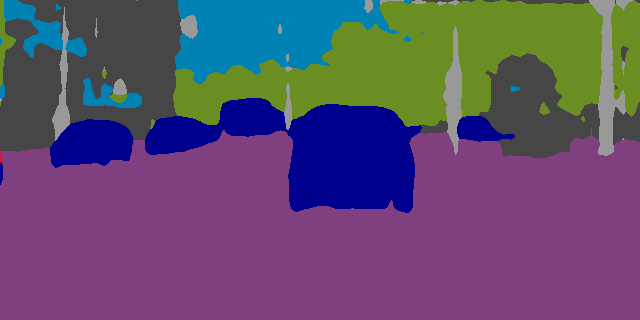} & \includegraphics[height=1.0cm]{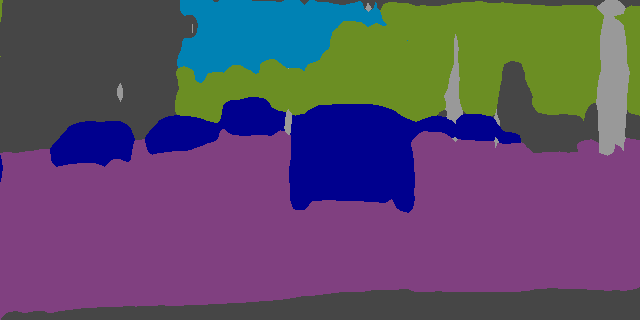} &
       \includegraphics[height=1.0cm]{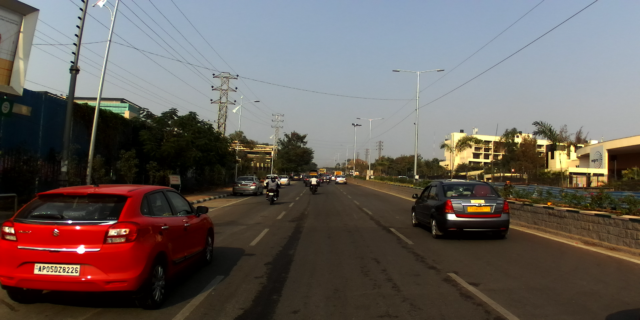} & \includegraphics[height=1.0cm]{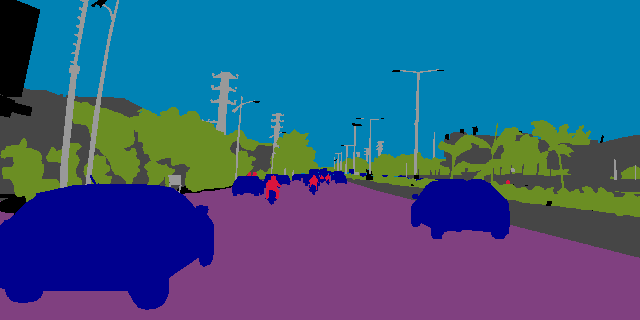} & \includegraphics[height=1.0cm]{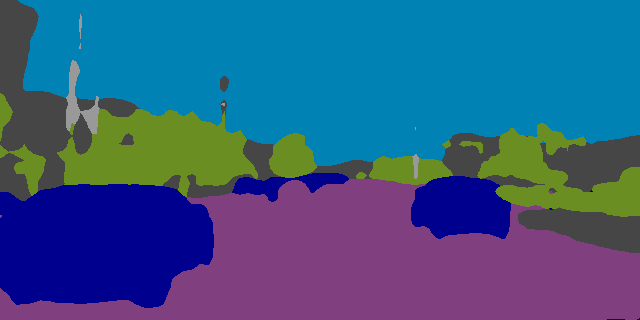} & \includegraphics[height=1.0cm]{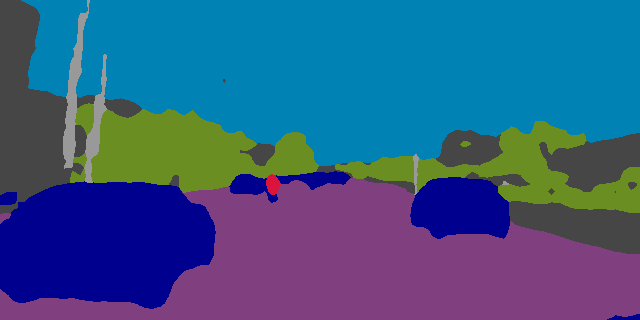} \\
       
       \includegraphics[height=1.0cm]{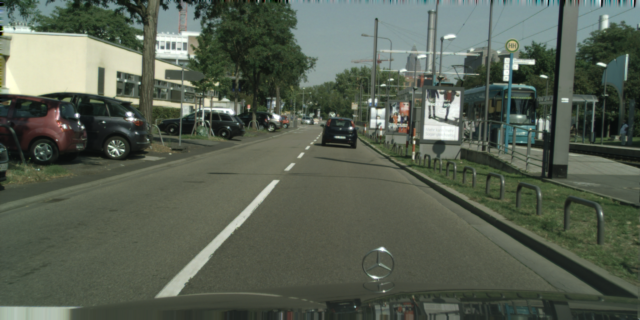} &
       \includegraphics[height=1.0cm]{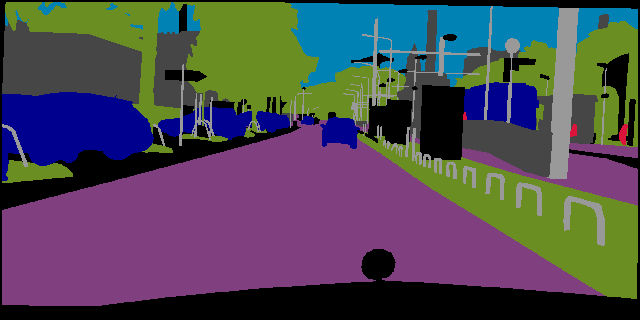} &
       \includegraphics[height=1.0cm]{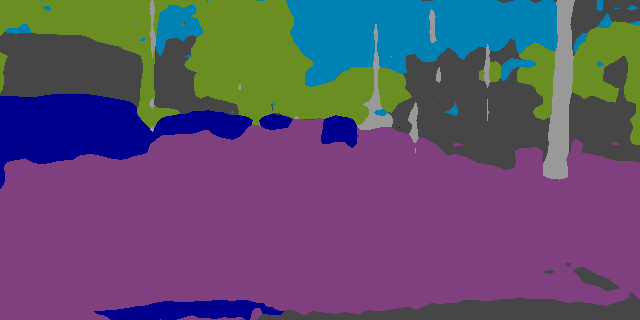} &
       \includegraphics[height=1.0cm]{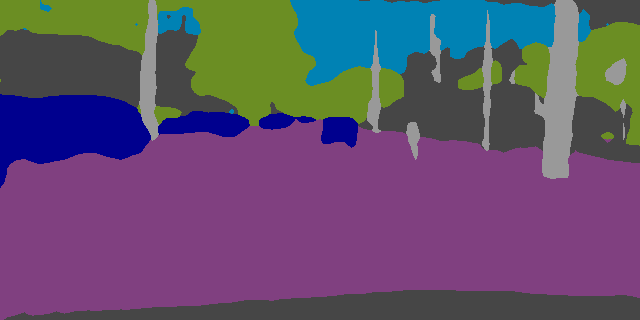} &
      \includegraphics[height=1.0cm]{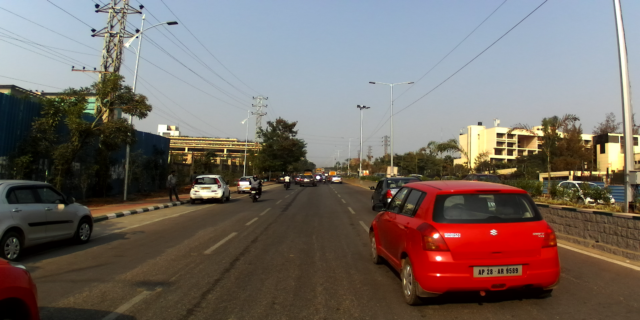} &
       \includegraphics[height=1.0cm]{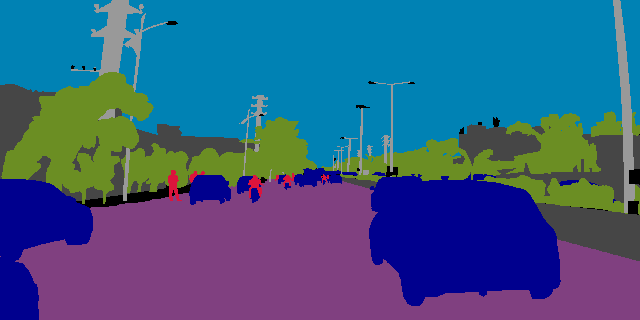} &
       \includegraphics[height=1.0cm]{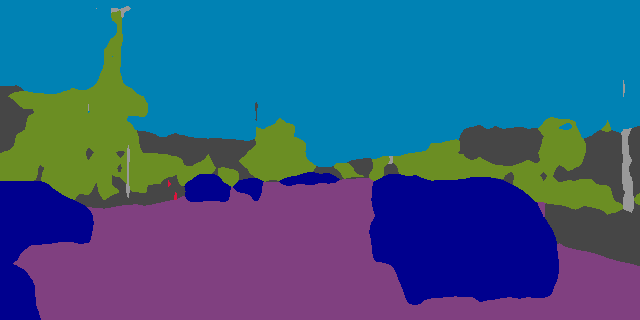} &
       \includegraphics[height=1.0cm]{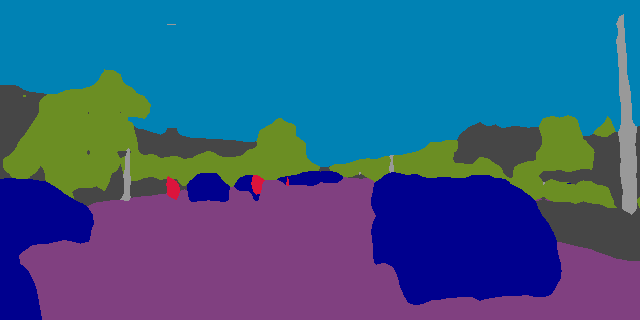}\\ 
       
       \includegraphics[height=1.0cm]{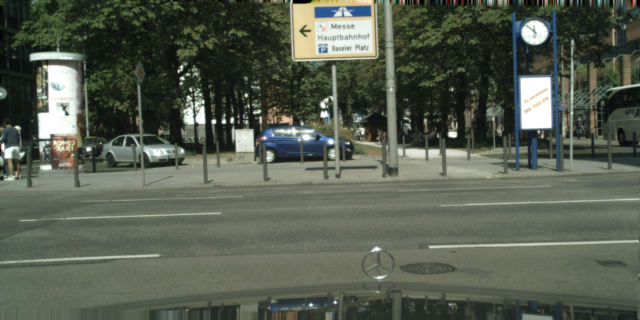} &
       \includegraphics[height=1.0cm]{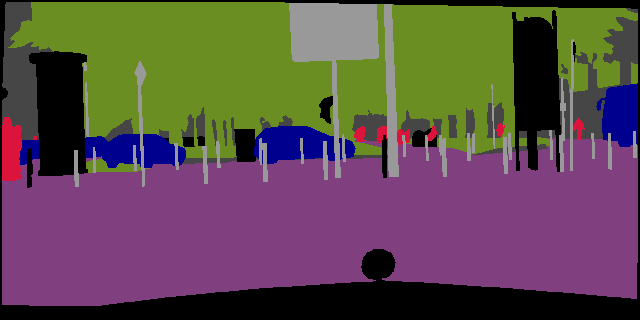} &
       \includegraphics[height=1.0cm]{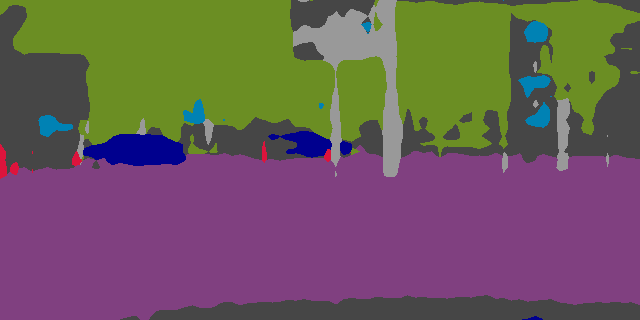} &
       \includegraphics[height=1.0cm]{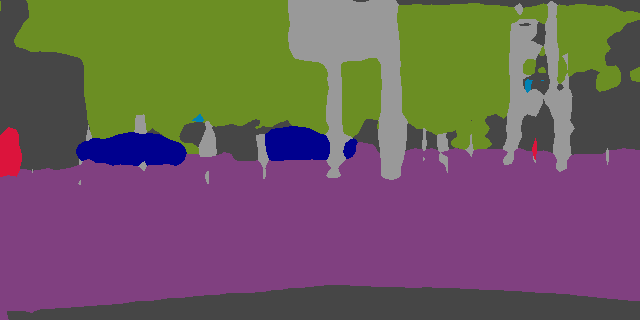} &
       \includegraphics[height=1.0cm]{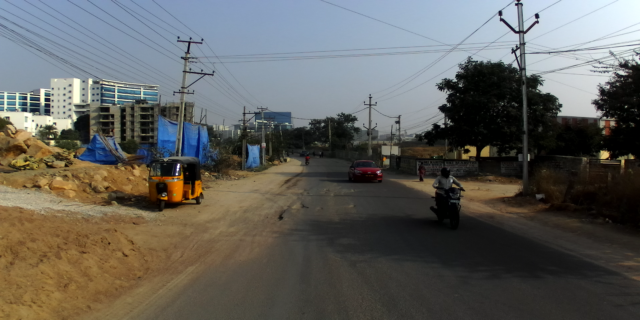} &
       \includegraphics[height=1.0cm]{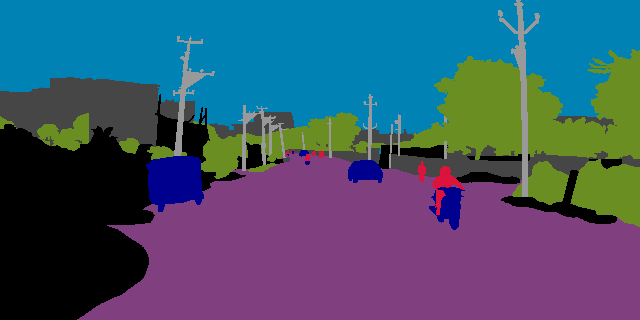} &
       \includegraphics[height=1.0cm]{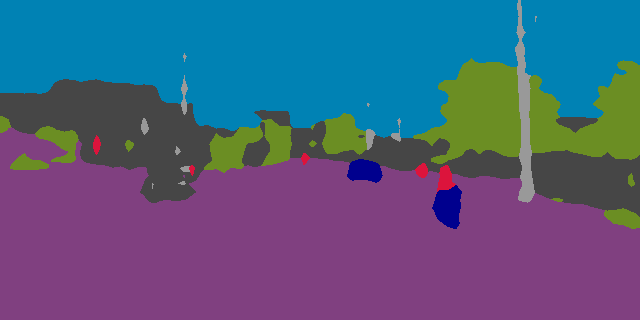} &
       \includegraphics[height=1.0cm]{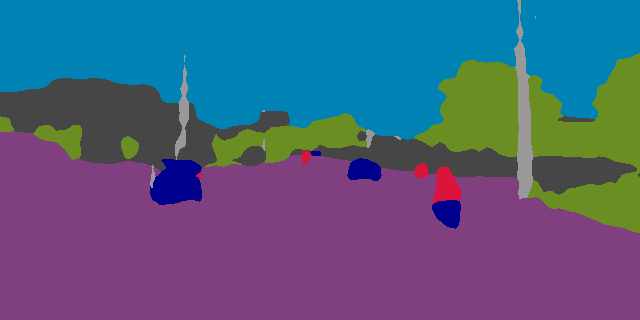}\\
       
       \includegraphics[height=1.0cm]{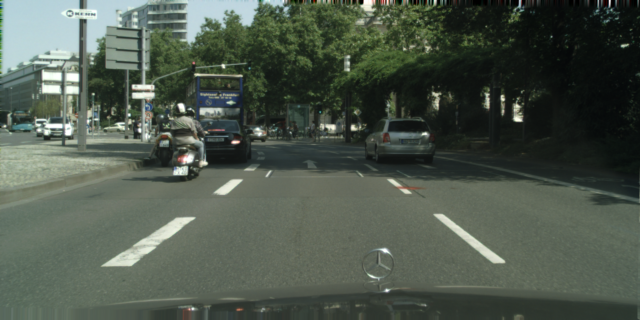} &
       \includegraphics[height=1.0cm]{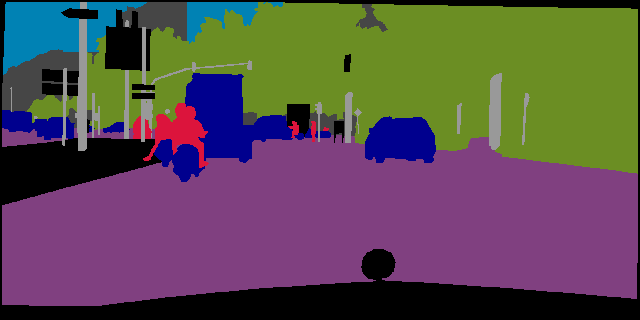} &
       \includegraphics[height=1.0cm]{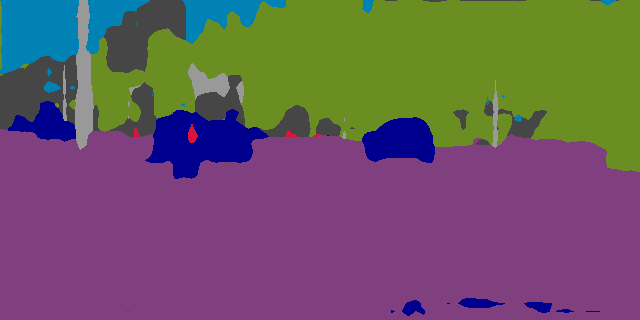} &
       \includegraphics[height=1.0cm]{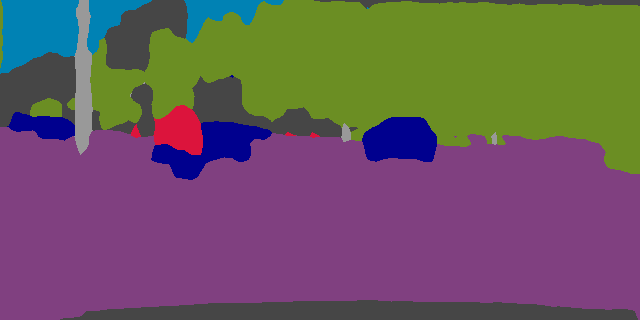} &
       \includegraphics[height=1.0cm]{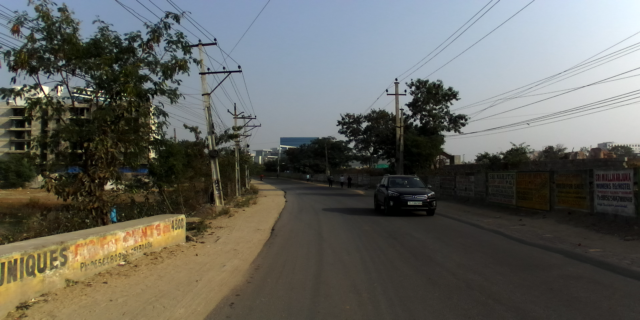} &
       \includegraphics[height=1.0cm]{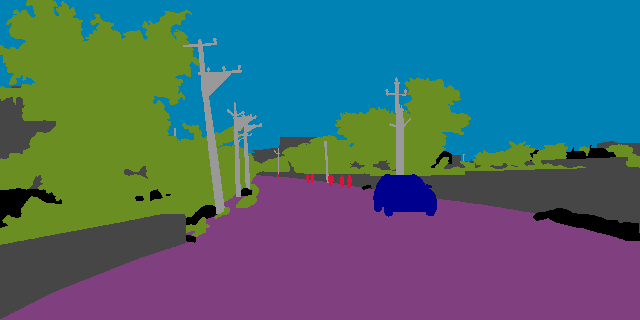} &
       \includegraphics[height=1.0cm]{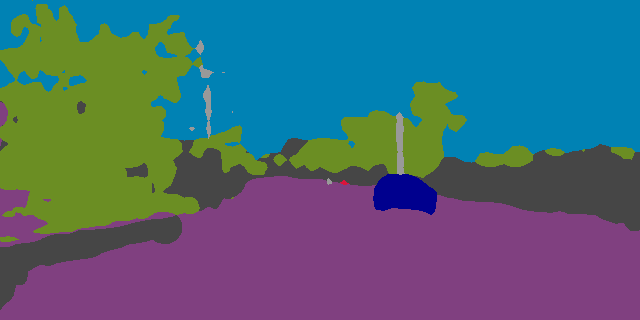} &
       \includegraphics[height=1.0cm]{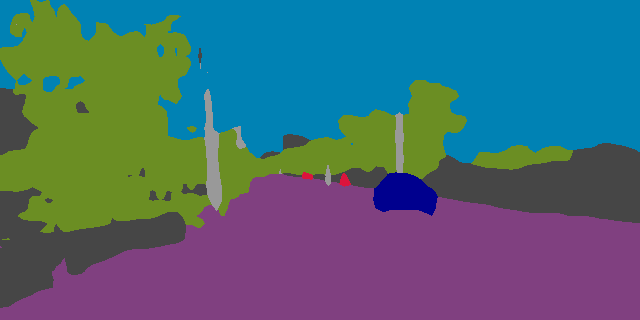}\\
       
       \includegraphics[height=1.0cm]{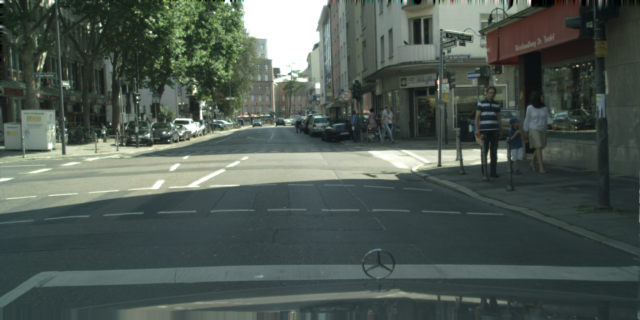} &
       \includegraphics[height=1.0cm]{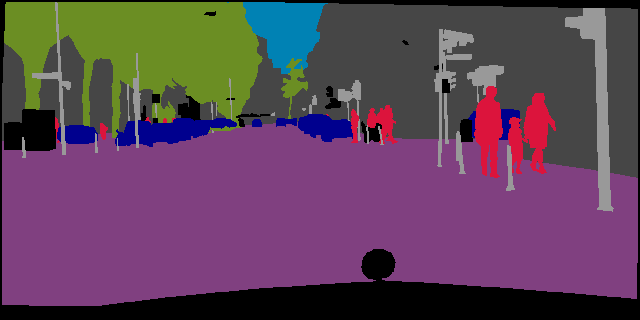} &
       \includegraphics[height=1.0cm]{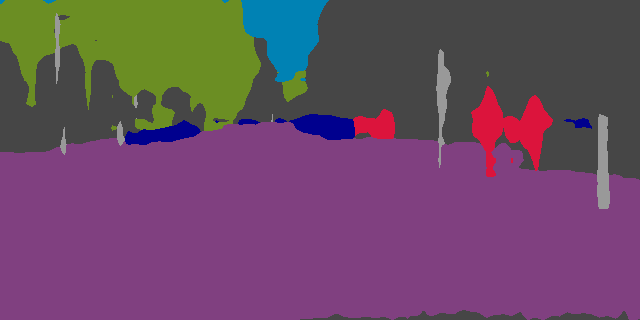} &
       \includegraphics[height=1.0cm]{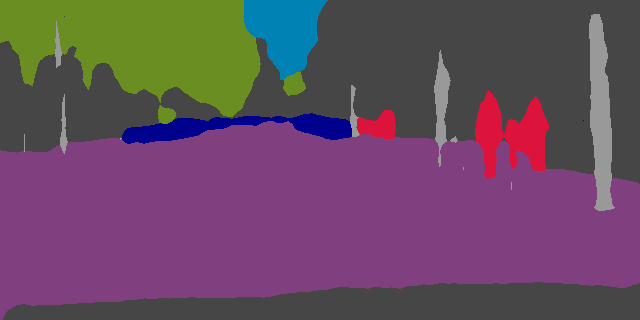} &
       \includegraphics[height=1.0cm]{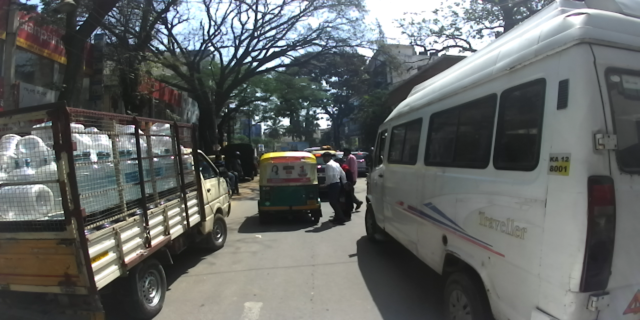} &
       \includegraphics[height=1.0cm]{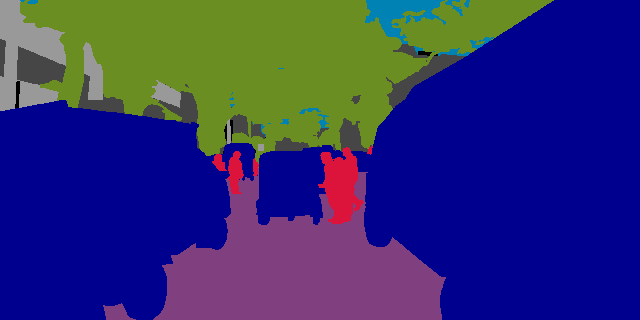} &
       \includegraphics[height=1.0cm]{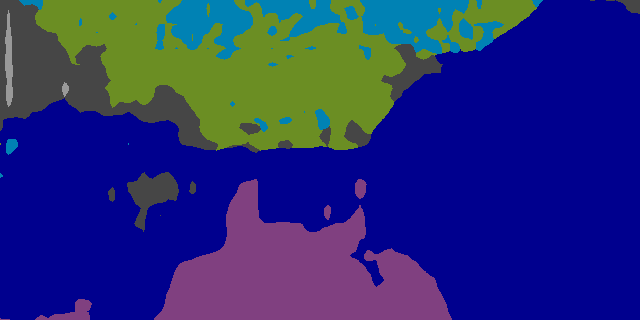} &
       \includegraphics[height=1.0cm]{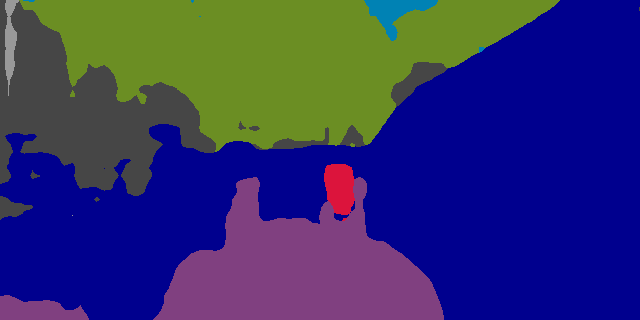}\\

    \end{tabular}
    }\vspace{-0.4cm}
    \caption{Qualitative comparison with \emph{MTKT} on the \emph{7-classes} benchmark and in the GTA5 to Cityscapes and IDD setting.}\vspace{-0.4cm}
    
    
    \label{fig:visualization}%
\end{figure*}

\noindent\textbf{Qualitative Comparison.}
Fig.~\ref{fig:visualization} shows a qualitative comparison with \emph{MTKT} on the \emph{7-classes} benchmark when adapting from GTA5 to Cityscapes and IDD. From these visualizations, we can see that \emph{\ours} segment better small objects such as 'human'  or 'object' classes. This difference is especially clear on the IDD dataset. 

\noindent\textbf{Summary of all the Settings.}
To complete this evaluation in the \emph{Synthetic to Real} scenario, we report in Tab.~\ref{table:sotaSummary} the average mIoU considering all the possible target configurations on the \emph{19-classes} benchmark. Results on the \emph{7-classes} benchmark are reported in supplement. For the \emph{19-classes} benchmarks, the proposed method is compared with the best respective competitor. In short, we observe that \emph{\ours} obtains the best performance in all configurations and on all the domains. These experiments demonstrate the robustness of our approach.

\setlength{\tabcolsep}{4pt}
\begin{table}[h]
\def\arraystretch{.9}
\begin{center}

\resizebox{0.45\textwidth}{!}{
\begin{tabular}{c c c|c| c c c |c}
\hline
 \multicolumn{3}{c|}{\bf Target} &  \multirow{2}*{\bf method} & \multicolumn{3}{c|}{\bf mIoU} & \bf mIoU\\ 
 C & I & M & ~ & C & I & M & Avg.\\ \hline

 \multirow{3}*{$\surd$} & \multirow{3}*{$\surd$} & \multirow{3}*{-} & \emph{CCL}~\cite{isobe2021multi} & 45.0 & 46.0 & - & 45.5 \\ 
 ~ & ~ & ~ & \emph{ADAS}~\cite{lee2022adas} & 45.8 & 46.3 & - &  46.1 \\
 ~ & ~ & ~ & \ours (Ours) & \textbf{47.1} & \textbf{49.3} & - & \bf 48.2 \\ \hline
  \multirow{3}*{$\surd$} & \multirow{3}*{-} & \multirow{3}*{$\surd$} & \emph{CCL}~\cite{isobe2021multi} & 45.1 & - & 48.8 & 47.0 \\ 
 ~ & ~ & ~ & \emph{ADAS}~\cite{lee2022adas} & 45.8 & - & 49.2 &  47.5 \\
 ~ & ~ & ~ & \ours (Ours) & \textbf{47.9} & - & \textbf{51.8} & \bf 49.9 \\ \hline
  \multirow{3}*{-} & \multirow{3}*{$\surd$} & \multirow{3}*{$\surd$} & \emph{CCL}~\cite{isobe2021multi} & - & 44.5 & 46.4 & 45.5 \\ 
 ~ & ~ & ~ & \emph{ADAS}~\cite{lee2022adas} & - & 46.1 & 47.6 &  46.9 \\
 ~ & ~ & ~ & \ours (Ours) & - & \textbf{49.5} & \textbf{51.6} & \bf 50.6 \\ \hline
  \multirow{3}*{$\surd$} & \multirow{3}*{$\surd$} & \multirow{3}*{$\surd$} & \emph{CCL}~\cite{isobe2021multi} & 46.7 & 47.0 & 49.9 & 47.9 \\ 
 ~ & ~ & ~ & \emph{ADAS}~\cite{lee2022adas} & 46.9 & 47.7 & 51.1 &  48.6 \\
 ~ & ~ & ~ & \ours (Ours) & \textbf{47.2} & \textbf{48.7} & \textbf{51.4} & \bf 49.1 \\ \hline
\end{tabular}}
\end{center}\vspace{-0.4cm}
\caption{Summary of performances obtained on the \emph{19-classes} benchmark. Cityscapes, IDD and Mapillary are referred to as C, I and M respectively. We report the mIoU averaged over the target domains.}
\label{table:sotaSummary}
\end{table}
\setlength{\tabcolsep}{1.4pt}

\setlength{\tabcolsep}{1.4pt}

\setlength{\tabcolsep}{4pt}
\begin{table*}[h]
\def\arraystretch{.9}
\parbox{.6\linewidth}{
\centering

    \resizebox{0.6\textwidth}{!}{\begin{tabular}{c|ccc  c c |c c|c}
    
    \hline
    Model & Adv. & Self-Tr. &\emph{CrossDoNorm} &  $\mathcal{L}_{cst}$ & Rec. & C & I & Avg.\\ 
    \hline
    \emph{MTKT*} \cite{saporta2021mtaf}   & $\surd$  & &  &  &&  67.3   & 64.3 & 65.8\\
    (i)  &   & $\surd$ &  &   &   & 65.6   & 63.2 & 64.4\\ 
    \hline
    (ii)  &&$\surd$    & $\surd$  &   &  & 69.2  & 67.4 & 68.4 \\ 
    (iii)  & &$\surd$   & $\surd$ &  $\surd$   &  &  70.2   & 67.5 & 68.9 \\ 
    (iv)  &&$\surd$    & $\surd$  & & $\surd$ & 72.1 & 69.9 & 71.0 \\ 
    \hline
    (v) & &$\surd$    & $\surd$  & $\surd$ & $\surd$ & 72.6  & 70.0 & 71.3\\ 
    \hline
    
%
    \end{tabular}}
    \caption{Ablation study of the proposed method on \emph{7-classes} benchmark, in GTA5 $\rightarrow$ Cityscapes + IDD  configuration.}
    \label{tab:ablation1}
}
\hspace{0.01\textwidth}
\parbox{.34\linewidth}{
\centering

 \resizebox{0.34\textwidth}{!}{\begin{tabular}{c|c c c}
    
    \hline
    Rectification &   C & I & Avg.\\ 
    \hline
    Without &  70.2   & 67.5 & 68.9 \\
    Drop-Out-based \cite{gal2016dropout}&  70.7   & 68.7  & 69.7\\ 
    Auxiliary network \cite{zheng2021rectifying}  &   69.8   & 68.7 & 69.3\\
    Cooperative (ours)   &  \bf 72.6  & \bf 70.0 & \bf 71.3\\   
    \hline

    \end{tabular}}
    \caption{Ablation study on the GTA5 $\rightarrow$ Cityscapes + IDD  configuration: Rectification strategy}
    \label{tab:ablation2}
}
\end{table*}

\setlength{\tabcolsep}{1pt}

\subsection{Ablation Study}
To illustrate the impact of the proposed cooperative self-training and rectification, we present a detailed ablation study. We present several variants of \emph{\ours}. First, we employ our architecture but with the adversarial training scheme of \emph{MTKT} \cite{saporta2021mtaf}. The goal of this variant is to show that our performance gain is not due to our slight architectural change in the classifiers. This variant is referred to as \emph{MTKT*}. Then we ablate different parts of our model: (i) uses a simple Self-training with pseudo-labeling without cross domain interactions. (ii) performs style-transfer and employs the cross-domain pseudo-label loss $\mathcal{L}_{pl}^{sty}$ in Eq.~\eqref{eqn:pl_cross}.  (iii) adds the consistency loss given in Eq.~\eqref{eqn:cst}. (iv) employs our rectified loss but does not use the consistency loss. Finally, (v) denotes our full models.  

The lower performance of \emph{MTKT*} demonstrates that the higher performance of \emph{\ours} is not due to the use of a different classifier. Then, we can observe that (i) under-performs \emph{MTKT}* showing that naively replacing adversarial training by self-training does not work well. Adding \emph{CrossDoNorm} in (ii) results in a $4\%$ gain that is further increased when a consistency loss is added (see (iii)). Cooperation between domains can be also obtained by introducing cross-domain rectification (see (iv)) but the experiments show that combining both consistency and pseudo label rectification leads to the best performance.


To complete this ablation study, we evaluate different solutions to assess the rectification weights $w_i$ in the same setting as Tab.~\ref{tab:ablation1}. We consider different possibilities. Cooperative cross domain rectification can be replaced by the consistency between predictions obtained with multiple drop-out sampling \cite{gal2016dropout}. An auxiliary network can also be employed as in~\cite{zheng2021rectifying} to estimate the uncertainty. Average mIoUs are reported in Tab.~\ref{tab:ablation2} with these different rectification approaches. We observe our approach which benefits from the multiple target domains, achieves a $2.4\%$ gain, which demonstrates that leveraging the multiple target domains is essential to achieve robust pseudo-labeling.

\vspace{-3mm}
\section{Conclusion}
\vspace{-3mm}
We presented \emph{CoaST}, a new method for multi-target domain adaptation in semantic segmentation. We introduced a self-training strategy that uses pseudo-labels in conjunction with style-transfer to favor consistency between classifiers. Besides, we employed consistency between the predictions from the different classifiers as an uncertainty measure allowing better use of the pseudo-labels. We conducted experiments in two benchmarks and several settings and demonstrated that the proposed method outperforms state-of-the-art approaches. 

\noindent\textbf{Acknowledgements.}
This paper has been partially supported by the French National Research
Agency (ANR) in the framework of its Technological Research JCJC pro-
gram (Odace, project ANR-20-CE23-0027), NSFC (62176155), Shanghai Municipal Science and Technology Major Project (2021SHZDZX0102), the EU ISFP PROTECTOR
(101034216) project and the EU H2020 MARVEL (957337) project.

{\small
\bibliographystyle{ieee_fullname}
\bibliography{egbib}
}

\clearpage
\appendix
\vspace*{1em}{\centering\large\bf%
 Supplementary Material for \\  Cooperative Self-Training for Multi-Target \\ \hspace{2.8em}Adaptive Semantic Segmentation%
\vspace*{1.5em}}

\setcounter{table}{0}
\renewcommand{\thetable}{A\arabic{table}}%
\setcounter{figure}{0}
\renewcommand{\thefigure}{A\arabic{figure}}%
\setcounter{equation}{0}
\renewcommand{\theequation}{A\arabic{equation}}%

The supplementary material is organized as follows: Sec. \ref{app:notation} summarizes the notations used. Sec. \ref{sec:set_up_sup} describes the experimental details of our work. Sec. \ref{sec:app-ablation} reports the ablation study on hyperparameter sensitivity. Sec. \ref{sec:app-detailed-comparison} lists detailed quantitative comparisons on various configurations.   

\section{Notation}\label{app:notation}
We summarize in Table~\ref{tab:app-notations} the notation used throughout the paper:

\begin{table}[!h]
\centering

\small

    \begin{tabularx}{\columnwidth}{ll}
    \toprule
    Notation & Description \\
    \midrule
    $\data\tsource = \{(\vx\tsource_n, \vy\tsource_n)\}^{N\tsource}_{n=1}$ & Source data set \\
    $\data\ttarget{i} = \{\vx\ttarget{i}_n\}^{N\ttarget{i}}_{n=1}$ & $i$\ths target data set \\
    $\vx\tsource \in \mathbb{R}^{H \times W \times 3}$ & Source inputs \\
    $\vy\tsource \in \mathbb{R}^{H \times W \times K}$ & Source labels \\
    $\vx\ttarget{i} \in \mathbb{R}^{H \times W \times 3}$ & Inputs from $i$\ths target domain \\
    $f$ & Model function \\
    $\Phi$ & Encoder network \\
    $C\ttarget{i}$ & $i$\ths target domain-specific classifier  \\
    $C^A$ & Domain-agnostic classifier  \\
    $D\ttarget{i}$ & $i$\ths target domain-specific discriminator\\
    $\hat{\mpred}\ttarget{i} \in \mathbb{R}^{H \times W \times K}$ & \makecell[l]{Prediction from $i$\ths target \\ domain-specific classifier} \\
    $\mbf{e}_k$ & One-hot encoding operator\\
    $\hat{\vy}\ttarget{i} \in \mathbb{R}^{H \times W}$ & \makecell[l]{Pseudo-label for $i$\ths target domain \\ sample} \\
    $\vz\ttarget{i}_l = \Phi_l(\vx\ttarget{i})$ & \makecell[l]{Latent feature of $\vx\ttarget{i}$ at $l$\ths layer in \\ $\Phi$\ths layer}\\
    $\bm{\mu}\ttarget{i}, \bm{\sigma}\ttarget{i}$ & \makecell[l]{Style vectors  (channel-wise mean and \\ standard deviation) for $i$\ths target \\ domain input $\vx\ttarget{i}$}\\
    $\vz\ttarget{i \to j}_l$ & \makecell[l]{Latent stylized feature of $\vx\ttarget{i}$ with \\ content from $\vx\ttarget{i}$ and style from $\vx\ttarget{j}$} \\
    $w_i$ & \makecell[l]{Averaged rectification weight for the \\ sample $\vx\ttarget{i}$}\\
    \bottomrule
    \end{tabularx}
\caption{Notation used throughout the paper}
\label{tab:app-notations}
\end{table}

\section{Experimental Details}
\label{sec:set_up_sup}

\noindent\textbf{Datasets.} We evaluate our method on two benchmarks previously used in the literature. These benchmarks are based on four datasets:

\begin{itemize}
\setlength\itemsep{-0.3em}
    \item \emph{GTA5}~\cite{richter2016playing} is collected from the video game GTA5. The dataset contains 24966 labeled images in total where the image resolution is 1914 $\times$ 1052. The synthetic nature of this dataset makes it very relevant for domain adaptation experiments.
    \item \emph{Cityscapes}~\cite{cordts2016cityscapes} is a large-scale dataset that has 2975 training and 500 validation labeled images collected mainly in German cities. 
    \item \emph{Mapillary}~\cite{neuhold2017mapillary} contains 18000 training and 2000 validation high-resolution images collected from all over the world. Compared to Cityscapes, this dataset is more diverse.
    \item \emph{IDD}~\cite{varma2019idd} is collected on Indian roads and it has 6993 and 981 finely annotated images in training and validation sets respectively. \emph{IDD} is very challenging since India cities visually differ from the cities depicted in the other datasets.
\end{itemize}

\noindent\textbf{Implementation details.} In the warm-up stage, we employ the hyper-parameters as~\cite{saporta2021mtaf} except that we extend the warm-up stage from 20K to 60K iterations to get better initial pseudo-labels for the self-training stage. In the second stage, we use Stochastic Gradient Descent optimizer with learning rate $1.0\times10^{-4}$ to train the model for another 60K iterations. In all the experiments in the \emph{7-classes} and \emph{19-classes} settings, we use random crop of size $320 \times 160$ and $512 \times 256$ respectively to accelerate the training. In the second stage, we use strong data-augmentation and update the pseudo-labels every 10K iterations.

\section{Ablation Study of Hyperparameters}
\label{sec:app-ablation}
In the final objective of our proposed \ours, we weigh all the constituent losses and set other hyperparameters with a value that equals to 1. This disposes off the need to have a target validation set, which indeed is not available for any UDA setting. Nevertheless, below we study the sensitivity of \ours with respect to two hyperparameters over ranges of possible values.

\noindent\textbf{Ratio of Pair-wise Losses}. We perform an ablation study on the weighing hyperparameter $\lambda$ that weighs the pair-wise losses: consistency loss $\mathcal{L}_\mathrm{cst}$ and the rectified segmentation losses $\bar{\mathcal{L}}^\mathrm{sty}_\mathrm{pl}$. The weighted training objective of our \ours, first introduced in Eqn.~\ref{eqn:final-objective} of the main paper, is written as:

\vspace{-5mm}
\begin{align}
\label{eqn:app-final-objective}
\begin{split}
  \mathcal{L}_\mathrm{\emph{\ours}}=\sum_{(\vx\tsource, \vy\tsource) \in \data\tsource} \mathcal{L}_\mathrm{seg}(\vx\tsource, \vy\tsource) \\  + \sum_{i=1}^M \sum_{\vx\ttarget{i}\in \data\ttarget{i}} \Big[ \frac{1}{M}\mathcal{L}_\mathrm{kd}(\vx\ttarget{i}) +\bar{\mathcal{L}}_\mathrm{pl}(\vx\ttarget{i}) \\
  +{\lambda}\frac{1}{M-1}\sum_{\substack{j=1\\j\neq i}}^M\sum_{\vx\ttarget{j}\in \data\ttarget{j}}
  \big(\bar{\mathcal{L}}_\mathrm{pl}^\mathrm{sty}(\vx\ttarget{i},\vx\ttarget{j}) \\ +  \mathcal{L}_\mathrm{cst}(\vx\ttarget{i},\vx\ttarget{j})\big)\Big] 
  \end{split} 
\end{align}

\begin{figure*}%
    \centering
    \resizebox{0.9\textwidth}{!}{
    \begin{tabular}{cc}
       \includegraphics[width=0.5\textwidth]{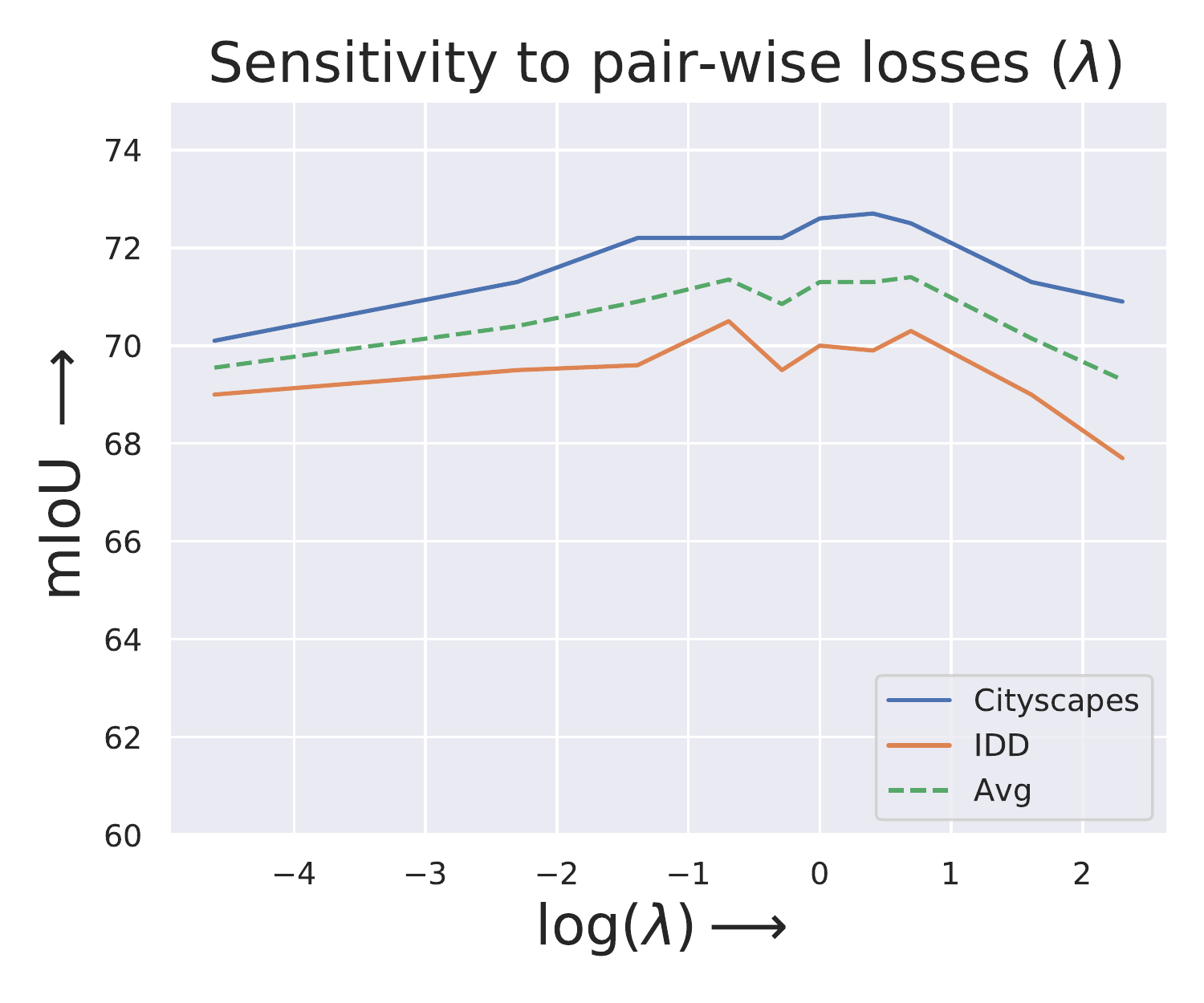} & \includegraphics[width=0.5\textwidth]{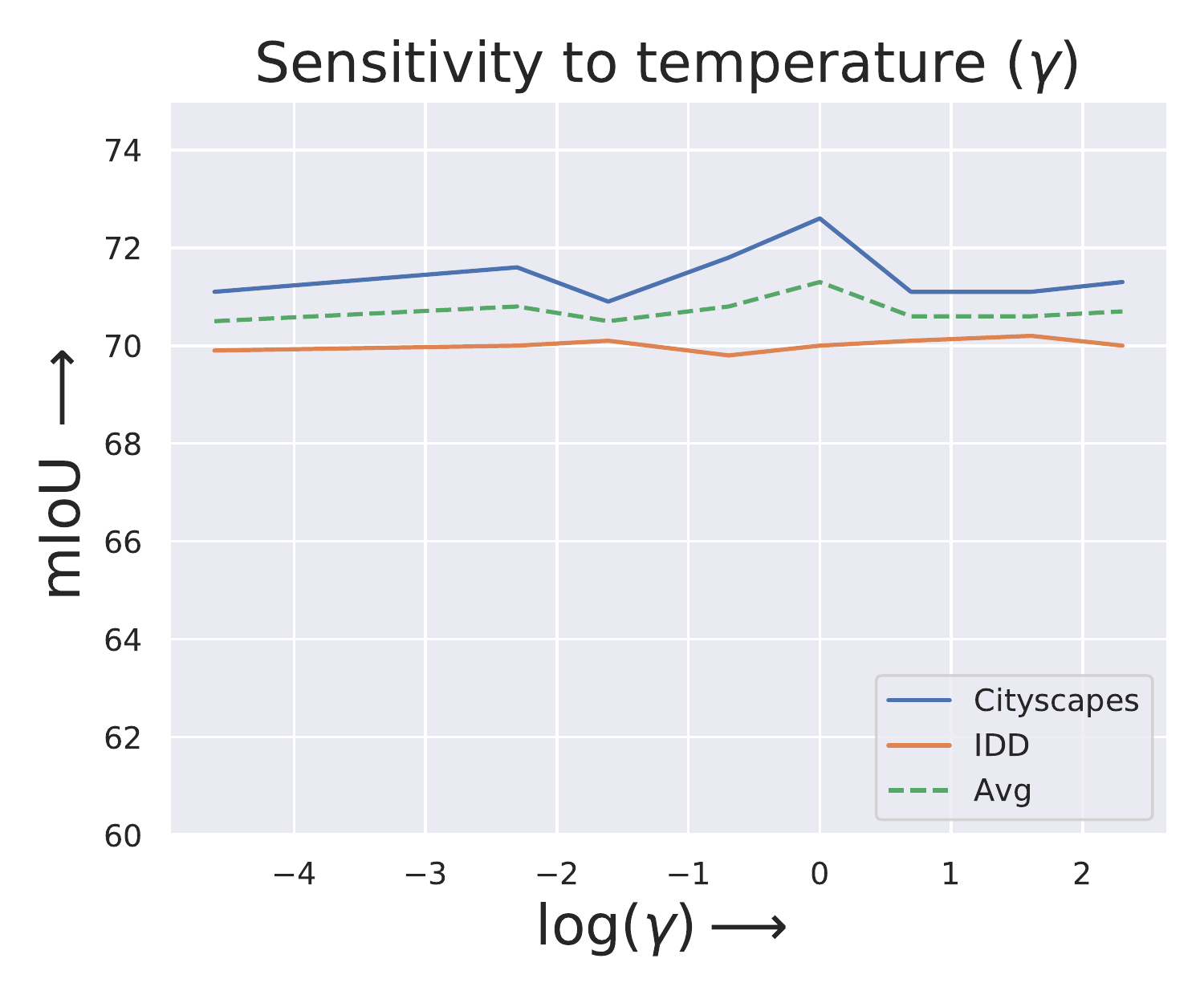} \\
       
    \end{tabular}
    }
    \caption{Sensitivity analysis of our proposed \ours for the \emph{7-class} MTDA configuration of GTA5 $\rightarrow$ Cityscapes + IDD. \textbf{Left}: we vary the pair-wise loss weight $\lambda$ and evaluate the mIoU for the target domains. The performance curve remains stable over a wide operating window, and starts to degrade only for extreme values of the $\lambda$. \textbf{Right}: we vary the temperature $\gamma$ and evaluate the mIoU for the target domains. We notice that the average mIoU varies slightly with $\gamma$.  On the x-axis we plot the logarithmic values of the hyperparameters for clarity}

    \label{fig:app-sensitivity}%
\end{figure*}

From the Fig.~\ref{fig:app-sensitivity} (left), we can see that, the mIoU remains fairly stable over a wide operating window of $\lambda$. The performance starts to drop only when we increase the value of $\lambda$ to large values. This is reasonable because when $\lambda=10$, the $\mathcal{L}_\mathrm{cst}$ and the $\bar{\mathcal{L}}^\mathrm{sty}_\mathrm{pl}$ starts to dominate the other losses in Eqn~\ref{eqn:app-final-objective}. We observe a well-behaved training dynamics when we set the value of $\lambda$ to standard value of 1, or $\log \lambda = 0$.

\noindent\textbf{Temperature}. The rectification weight described in the Eqn.~\ref{eqn:rectification-weight} of the main paper is obtained by applying an exponential operation on the consistency score. To recap, the exp(.) function is used to bound the KL-divergence consistency score between ]0,1], which otherwise is unbounded. The rectification weight can be regulated by using a \textit{temperature} hyperparameter $\gamma$, that controls the steepness of the exp(.) curve. In other words, higher the value of $\gamma$, more quickly the curve goes to zero, and vice-versa. The rectification weight which is a function of $\gamma$ is given as:

\begin{equation}
  w_i = \frac{1}{M-1}  \sum_{j=1, j\neq i}^M \exp{(-{\gamma} \mathcal{L}_\mathrm{kl}(\hat{\mpred}\ttarget{i}, \hat{\mpred}\ttarget{i \to j})})
\end{equation}

%
%
%

It can be observed from the Fig.~\ref{fig:app-sensitivity} (right) that the performance of \ours does not vary much while changing the temperature $\gamma$. Indeed, we see that the average mIoU remains in a tight range of 70.5\% to 71.3\%, even for extreme values of $\gamma$. Note that we vary the value of $\gamma$ between 0.01 and 10 in our ablation study, whereas we report the logarithmic values of $\gamma$ in Fig.~\ref{fig:app-sensitivity} on the x-axis for clarity of the plot.

%
%
%

\setlength{\tabcolsep}{1.4pt}
\begin{table*}[!h]
\begin{center}

\resizebox{0.9\textwidth}{!}{
    \begin{tabular}{c|c|c c c c c c c |c|c}
    \hline
    \noalign{\smallskip}
    \multicolumn{11}{c}{\bf GTA5 $\rightarrow$ Cityscapes + Mapillary}\\ 
    \noalign{\smallskip}
    \hline
    \bf Method & \bf Target & \bf flat & \bf constr & \bf object & \bf nature & \bf sky & \bf human & \bf vehicle & \bf mIoU & \bf Avg. \\ \hline
    
    \multirow{2}*{\emph{Individual}~\cite{vu2018advent}} & C & 93.5 & 80.5 & 26.0 & 78.5 & 78.5 & 55.1 & 76.4 & 69.8 & \multirow{2}*{69.7} \\  
    
    ~ & M & 89.5 & 72.6 & 31.0 & 75.3 & 94.1 & 50.7 & 73.8 & 69.6 & ~ \\ \hline

    \multirow{2}*{\emph{Data Comb.}~\cite{vu2018advent}} & C & 93.1 & 80.5 & 24.0 & 77.9 & 81.0 & 52.5 & 75.0 & 69.1 & \multirow{2}*{68.9} \\  
    ~ & M & 90.0 & 71.3 & 31.1 & 73.0 & 92.6 & 46.6 & 76.6 & 68.7 & ~ \\ \hline
    
    \multirow{2}*{\emph{Multi-Dis}~\cite{saporta2021mtaf}} & C & 94.5 & 80.8 & 22.2 & 79.2 & 82.1 & 47.0 & 79.0 & 69.3 & \multirow{2}*{69.5} \\  
    ~ & M & 89.4 & 71.2 & 29.5 & \bf 76.2 & 93.6 & 50.4 & 78.3 & 69.8 & ~ \\ \hline
    
    \multirow{2}*{\emph{MTKT}~\cite{saporta2021mtaf}} & C & 95.0 & 81.6 & 23.6 & 80.1 & \bf 83.6 & 53.7 & 79.8 & 71.1 & \multirow{2}*{70.9} \\  
    ~ & M & \bf 90.6 & 73.3 & 31.0 & 75.3 & \bf 94.5 & 52.2 & 79.8 & 70.8 & ~ \\ \hline
    
    \multirow{2}*{\emph{ADAS}~\cite{lee2022adas}($1024\times512$)} & C & \bf 96.4 & 83.5 & \bf 35.1 & \bf 83.6 & \bf 84.9 & \bf 62.3 & \bf 81.3 & \bf 75.3 & \multirow{2}*{\bf 73.9} \\  
    ~ & M & 88.6 & 73.7 & \bf 41.0 & \bf 75.4 & \bf 93.4 & \bf 58.5 & 77.2 & \bf 72.6 & ~ \\ \hline
    
    \multirow{2}*{\emph{\ours} (Ours)} & C & 94.7 & \bf 84.4 & 29.3 & 81.6 & 77.7 & 57.1 & \bf 81.3 & 72.3 & \multirow{2}*{72.3} \\  
    ~ & M & 89.2 & \bf 74.9 & 37.5 & 74.6 & 89.2 & 57.9 & \bf 82.8 & 72.3 & ~ \\ \hline
    \end{tabular}
}
\end{center}
\caption{The comparison of \ours with the state-of-the-art on the \emph{7-classes} benchmark using the GTA5 $\rightarrow$ Cityscapes + Mapillary  configuration. We observe that \ours outperforms MTKT on several classes and also on average}
\label{table:G2CM}
\end{table*}
\setlength{\tabcolsep}{1.4pt}

\section{Quantitative Comparison.}
\label{sec:app-detailed-comparison}

\subsection{Detailed Results of the Synthetic to Real Settings}

In the Tab.~\ref{table:sotaSummary} of the main paper, we reported the summary of the performances on all the settings with GTA5 as the source domain. In this section, we report the detailed class-wise results for those settings. The Tab.~\ref{table:G2CM},~\ref{table:G2MI} and~\ref{table:G2CMI} detail the results on the \emph{7-class} setting while the Tab.~\ref{table:detail19} detail the results on the  \emph{19-class} setting. Note that the detailed results of G2CI are already shown in Tab.~\ref{table:G2CI} and the Tab.~\ref{table:G2CI19} of the main paper.

In Tab.~\ref{table:G2CM},~\ref{table:G2MI} and~\ref{table:G2CMI}, we can see that our \ours outperforms all the baselines and \emph{MTKT}~\cite{saporta2021mtaf} in \emph{7-class} benchmark for most of the classes. These results are in-line with the summarized results reported in the main paper and confirm the consistent gain provided by our \ours for the majority of the classes. In Tab.~\ref{table:detail19}, we show the detailed comparison with \emph{Individual} and \emph{MTKT} in \emph{19-class} benchmark. Note that, the detailed comparison with scores reported for every class is not reported in paper introducing \emph{CCL}~\cite{isobe2021multi} and \emph{ADAS}~\cite{lee2022adas}. Since their codes are not publicly available, we could not provide the detailed class-wise scores. The comparison with \emph{CCL}~\cite{isobe2021multi} and \emph{ADAS}~\cite{lee2022adas} could only be reported as in Tab.~\ref{table:sotaSummary} of the main paper.

\setlength{\tabcolsep}{1.4pt}
\begin{table*}[!t]
\begin{center}

\resizebox{0.9\textwidth}{!}{
    \begin{tabular}{c|c|c c c c c c c |c|c}
    \hline
    \noalign{\smallskip}
    \multicolumn{11}{c}{\bf GTA5 $\rightarrow$ Mapillary + IDD}\\ 
    \noalign{\smallskip}
    \hline
    \bf Method & \bf Target & \bf flat & \bf constr & \bf object & \bf nature & \bf sky & \bf human & \bf vehicle & \bf mIoU & \bf Avg. \\ \hline
    
    \multirow{2}*{\emph{Individual}~\cite{vu2018advent}} & M & 89.5 & 72.6 & 31.0 & 75.3 & 94.1 & 50.7 & 73.8 & 69.6 & \multirow{2}*{67.4} \\  
    
    ~ & I & 91.2 & 53.1 & 16.0 & 78.2 & 90.7 & 47.9 & 78.9 & 65.1 & ~ \\ \hline

    \multirow{2}*{\emph{Data Comb.}~\cite{vu2018advent}} & M & 89.6 & 71.0 & \bf 34.2 & 74.5 & 92.9 & 47.3 & 78.6 & 69.7 & \multirow{2}*{67.9} \\  
    ~ & I & 91.8 & 54.0 & \bf 17.4 & 76.9 & 92.3 & 51.4 & 78.4 & 66.0 & ~ \\ \hline
    
    \multirow{2}*{\emph{Multi-Dis}~\cite{saporta2021mtaf}} & M & \bf 89.9 & 71.7 & 28.7 & 76.0 & 93.6 & 51.6 & 79.7 & 70.2 & \multirow{2}*{68.1} \\  
    ~ & I & 91.4 & 54.9 & 14.6 & 78.5 & \bf 93.0 & 51.1 & 79.0 & 66.1 & ~ \\ \hline
    
    \multirow{2}*{\emph{MTKT}~\cite{saporta2021mtaf}} & M & 88.8 & 73.2 & 31.5 & 74.7 & \bf 94.1 & 52.5 & \bf 79.9 & 70.7 & \multirow{2}*{68.3} \\  
    ~ & I & 91.4 & 55.9 & 13.5 & 76.7 & 92.1 & \bf 52.3 & 79.4 & 65.9 & ~ \\ \hline
    
    \multirow{2}*{\emph{\ours} (Ours)} & M & \bf 90.5 & \bf 75.9 & 37.2 & \bf 73.6 & 90.8 & \bf 57.5 & 81.3 & \bf 72.4 & \multirow{2}*{\bf 70.6} \\  
    ~ & I & \bf 93.3 & \bf 60.9 & 19.8 & \bf 79.3 & 91.2 & 54.1 & \bf 82.6 & \bf 68.7 & ~ \\ \hline
    \end{tabular}
}
\caption{The comparison of \ours with the state-of-the-art on the \emph{7-classes} benchmark using the GTA5 $\rightarrow$ Mapillary + IDD  configuration. We observe that \ours outperforms MTKT on several classes and also on average}
\label{table:G2MI}
\end{center}
\end{table*}
\setlength{\tabcolsep}{1.4pt}

\setlength{\tabcolsep}{1.4pt}
\begin{table*}[!t]
\begin{center}

\resizebox{0.9\textwidth}{!}{
    \begin{tabular}{c|c|c c c c c c c |c|c}
    \hline
    \noalign{\smallskip}
    \multicolumn{11}{c}{\bf GTA5 $\rightarrow$ Cityscapes + Mapillary + IDD}\\ 
    \noalign{\smallskip}
    \hline
    \bf Method & \bf Target & \bf flat & \bf constr & \bf object & \bf nature & \bf sky & \bf human & \bf vehicle & \bf mIoU & \bf Avg. \\ \hline
    
    \multirow{3}*{\emph{Individual}~\cite{vu2018advent}} & C & 93.5 & 80.5 & 26.0 & 78.5 & 78.5 & 55.1 & 76.4 & 69.8 & \multirow{3}*{68.2} \\  

    ~ & M & 89.5 & 72.6 & 31.0 & 75.3 & 94.1 & 50.7 & 73.8 & 69.6 & ~ \\ 

    ~ & I & 91.2 & 53.1 & 16.0 & 78.2 & 90.7 & 47.9 & 78.9 & 65.1 & ~ \\ \hline

    \multirow{3}*{\emph{Data Comb.}~\cite{vu2018advent}} & C & 93.6 & 80.6 & 26.4 & 78.1 & 81.5 & 51.9 & 76.4 & 69.8 & \multirow{3}*{67.8} \\  
    ~ & M & 89.2 & 72.4 & 32.4 & 73.0 & 92.7 & 41.6 & 74.9 & 68.0 & ~ \\ 
    ~ & I & 92.0 & 54.6 & 15.7 & 77.2 & 90.5 & 50.8 & 78.6 & 65.6 & ~ \\ \hline

    \multirow{3}*{\emph{Multi-Dis}~\cite{saporta2021mtaf}} & C & \bf 94.6 & 80.0 & 20.6 & 79.3 & 84.1 & 44.6 & 78.2 & 68.8 & \multirow{3}*{68.2} \\  
    ~ & M & 89.0 & 72.5 & 29.3 & 75.5 & \bf 94.7 & 50.3 & 78.9 & 70.0 & ~ \\ 
    ~ & I & 91.6 & 54.2 & 13.1 & 78.4 & 93.1 & 49.6 & 80.3 & 65.8 & ~ \\ \hline
    
    \multirow{3}*{\emph{MTKT}~\cite{saporta2021mtaf}} & C & \bf 94.6 & \bf 80.7 & 23.8 & 79.0 & 84.5 & 51.0 & 79.2 & 70.4 & \multirow{3}*{69.1} \\  
    ~ & M & 90.5 & 73.7 & 32.5 & \bf 75.5 & 94.3 & 51.2 & \bf 80.2 & 71.1 & ~ \\ 
    ~ & I & 91.7 & 55.6 & 14.5 & 78.0 & \bf 92.6 & 49.8 & 79.4 & 65.9 & ~ \\ \hline
    
    \multirow{3}*{\emph{ADAS}~\cite{lee2022adas}($1024\times512$)} & C & \bf 95.8 & \bf 82.4 & \bf 38.3 & 82.4 & 85.0 & \bf 60.5 & 80.2 & \bf 74.9 & \multirow{3}*{71.3} \\  
    ~ & M & 89.2 & 71.5 & \bf 45.2 & \bf 75.8 & 92.3 & 56.1 & 75.4 & \bf 72.2 & ~ \\ 
    ~ & I & 89.9 & 52.7 & \bf 25.0 & 78.1 & 92.1 & 51.0 & 77.9 & 66.7 & ~ \\ \hline
    
    \multirow{3}*{\emph{\ours} (Ours)} & C & 94.4 & 80.2 & 27.0 & \bf 82.6 & \bf 88.3 & 54.6 & \bf 81.0 & 72.6 & \multirow{3}*{\bf 71.7} \\  
    ~ & M & \bf 91.7 & \bf 74.9 & 36.2 & 73.9 & 92.0 & \bf 57.5 & 79.5 & \bf 72.2 & ~ \\ 
    ~ & I & \bf 94.6 & \bf 62.0 & 21.0 & \bf 82.6 & \bf 92.6 & \bf 55.4 & \bf 83.7 & \bf 70.3 & ~ \\ \hline
    
    \end{tabular}
}
\caption{The comparison of \ours with the state-of-the-art on the \emph{7-classes} benchmark using the GTA5 $\rightarrow$ Cityscapes + Mapillary + IDD configuration. We observe that \ours outperforms MTKT on several classes and also on average. Particularly, the gain in performance for \ours over MTKT for the IDD is fairly substantial}
\label{table:G2CMI}
\end{center}
\end{table*}
\setlength{\tabcolsep}{1.4pt}

\setlength{\tabcolsep}{1.4pt}
\begin{table*}[!h]
\begin{center}

\resizebox{ \textwidth}{!}{
    \begin{tabular}{c|c|c c c c c c c c c c c c c c c c c c c|c|c}
    
    \hline
    \noalign{\smallskip}
    \multicolumn{23}{c}{\bf GTA5 $\rightarrow$ Cityscapes + IDD}\\
    \noalign{\smallskip}
    \hline
    \bf Method & \bf \rotatebox{90}{Target} & \bf\rotatebox{90}{road} & \bf\rotatebox{90}{sidewalk} & \bf\rotatebox{90}{building} & \bf\rotatebox{90}{walk} & \bf\rotatebox{90}{fence} & \bf\rotatebox{90}{pole} & \bf\rotatebox{90}{light} & \bf\rotatebox{90}{sign} & \bf\rotatebox{90}{veg} & \bf\rotatebox{90}{terrain} & \bf\rotatebox{90}{sky} & \bf\rotatebox{90}{person} & \bf\rotatebox{90}{rider} & \bf\rotatebox{90}{car} & \bf\rotatebox{90}{truck} & \bf\rotatebox{90}{bus} & \bf\rotatebox{90}{train} & \bf\rotatebox{90}{motor} & \bf\rotatebox{90}{bike} & \bf mIoU & \bf Avg. \\ \hline

    \multirow{2}*{\emph{Individual}~\cite{vu2018advent}} & C & \textbf{88.8} & 23.8 & \textbf{81.5} & 27.7 & 27.3 & 31.7 & 33.2 & 22.9 & 83.1 & 27.0 & 76.4 & 58.5 & \textbf{28.9} & 84.3 & \textbf{30.0} & 36.8 & \textbf{0.3} & 27.7 & \textbf{33.1} & 43.3 & \multirow{2}*{43.5} \\  
    ~ & I & \textbf{94.1} & 24.4 & \textbf{66.1} & 31.3 & 22.0 & 25.4 & 9.3 & 26.7 & 80.0 & \textbf{31.4} & \textbf{93.5} & 48.7 & 43.8 & 71.4 & 49.4 & 28.5 & \textbf{0} & 48.7 & \textbf{34.3} & 43.6 & ~ \\ \hline
    
    \multirow{2}*{\emph{MTKT}} & C & 88.5 & 37.2 & 79.1 & 22.8 & 19.8 & 26.3 & 33.8 & 16.7 & 84.8 & 34.2 & 80.5 & 54.9 & 15.0 & 84.1 & 27.5 & 41.2 & \textbf{0.3} & 27.9 & 7.0 & 41.7  & \multirow{2}*{42.5} \\  
    ~ & I & 82.7 & 24.2 & 54.2 & 29.3 & 22.0 & 24.8 & 8.4 & 52.0 & 78.7 & 18.2 & 92.1 & 38.6 & 51.0 & 72.5 & 60.8 & 27.6 & \textbf{0} & 54.5 & 14.1 & 43.3 & ~\\ \hline
    
    \multirow{2}*{\emph{\ours} (Ours)} & C & 81.7 & \textbf{38.3} & 71.0 & \textbf{33.3} & \textbf{30.7} & \textbf{35.1} & \textbf{38.2} & \textbf{37.6} & \textbf{86.4} & \textbf{46.9} & \textbf{81.9} & \textbf{63.4} & 27.4 & \textbf{84.5} & 29.4 & \textbf{45.6} & \textbf{0.3} & \textbf{32.6} & 31.3 & \textbf{47.1}  & \multirow{2}*{\textbf{48.2}} \\  
    ~ & I & 85.7 & \textbf{36.1} & 65.1 & \textbf{33.2} & \textbf{23.7} & \textbf{32.8} & \textbf{19.0} & \textbf{62.9} & \textbf{82.5} & 29.5 & 91.8 & \textbf{52.1} & \textbf{55.3} & \textbf{83.4} & \textbf{62.9} & \textbf{46.1} & \textbf{0} & \textbf{55.5} & 18.5 & \textbf{49.3} & ~\\ \hline

    \hline
    \noalign{\smallskip}
    \multicolumn{23}{c}{\bf GTA5 $\rightarrow$ Cityscapes + Mapillary}\\
    \noalign{\smallskip}
    \hline
    \bf Method & \bf \rotatebox{90}{Target} & \bf\rotatebox{90}{road} & \bf\rotatebox{90}{sidewalk} & \bf\rotatebox{90}{building} & \bf\rotatebox{90}{walk} & \bf\rotatebox{90}{fence} & \bf\rotatebox{90}{pole} & \bf\rotatebox{90}{light} & \bf\rotatebox{90}{sign} & \bf\rotatebox{90}{veg} & \bf\rotatebox{90}{terrain} & \bf\rotatebox{90}{sky} & \bf\rotatebox{90}{person} & \bf\rotatebox{90}{rider} & \bf\rotatebox{90}{car} & \bf\rotatebox{90}{truck} & \bf\rotatebox{90}{bus} & \bf\rotatebox{90}{train} & \bf\rotatebox{90}{motor} & \bf\rotatebox{90}{bike} & \bf mIoU & \bf Avg. \\ \hline

    \multirow{2}*{\emph{Individual}~\cite{vu2018advent}} & C & 88.8 & 23.8 & \textbf{81.5} & 27.7 & 27.3 & 31.7 & 33.2 & 22.9 & 83.1 & 27.0 & 76.4 & 58.5 & \textbf{28.9} & 84.3 & 30.0 & 36.8 & 0.3 & 27.7 & 33.1 & 43.3 & \multirow{2}*{44.7} \\  
    ~ & M & 81.1 & 18.6 & 74.8 & 23.9 & \textbf{28.9} & 30.3 & 35.7 & 33.7 & \textbf{78.4} & 40.7 & 93.3 & 49.5 & \textbf{42.3} & 80.4 & 35.1 & 34.2 & \textbf{17.8} & 41.8 & \textbf{36.1} & 46.1 & ~ \\ \hline
    
    \multirow{2}*{\emph{MTKT} \cite{saporta2021mtaf}} & C & \textbf{89.2} & 36.1 & \textbf{81.5} & 31.6 & 22.1 & 28.4 & 31.4 & 13.8 & 85.1 & 34.3 & \textbf{83.5} & 57.6 & 19.1 & \textbf{86.1} & \textbf{36.0} & 44.1 & 0.4 & \textbf{32.5} & 6.1 & 43.1  & \multirow{2}*{45.4} \\  
    ~ & M & \textbf{86.8} & 38.7 & 78.7 & 27.0 & 28.4 & 29.5 & 37.3 & 34.6 & \textbf{78.4} & 42.3 & \textbf{94.9} & 53.7 & 37.9 & \textbf{84.2} & 41.1 & 34.5 & 15.5 & \textbf{44.0} & 18.0 & 47.6 & ~\\ \hline
    
    \multirow{2}*{\emph{\ours} (Ours)} & C & 82.1 & \textbf{36.2} & 77.5 & \textbf{47.4} & \textbf{34.9} & \textbf{36.7} & \textbf{42.0} & \textbf{36.6} & \textbf{87.2} & \textbf{38.6} & 80.8 & \textbf{60.6} & 21.6 & \textbf{86.1} & 33.3 & \textbf{45.7} & \textbf{2.5} & 26.2 & \textbf{34.7} & \textbf{47.9}  & \multirow{2}*{\textbf{49.9}} \\  
    ~ & M & 84.7 & \textbf{44.4} & \textbf{80.3} & \textbf{35.7} & 27.7 & \textbf{37.2} & \textbf{45.1} & \textbf{51.8} & 73.8 & \textbf{42.4} & 93.7 & \textbf{64.5} & 42.2 & 83.9 & \textbf{49.0} & \textbf{44.0} & 10.0 & 38.5 & 35.6 & \textbf{51.8} & ~\\ \hline

    \hline
    \noalign{\smallskip}
    \multicolumn{23}{c}{\bf GTA5 $\rightarrow$ Mapillary + IDD}\\
    \noalign{\smallskip}
    \hline
    \bf Method & \bf \rotatebox{90}{Target} & \bf\rotatebox{90}{road} & \bf\rotatebox{90}{sidewalk} & \bf\rotatebox{90}{building} & \bf\rotatebox{90}{walk} & \bf\rotatebox{90}{fence} & \bf\rotatebox{90}{pole} & \bf\rotatebox{90}{light} & \bf\rotatebox{90}{sign} & \bf\rotatebox{90}{veg} & \bf\rotatebox{90}{terrain} & \bf\rotatebox{90}{sky} & \bf\rotatebox{90}{person} & \bf\rotatebox{90}{rider} & \bf\rotatebox{90}{car} & \bf\rotatebox{90}{truck} & \bf\rotatebox{90}{bus} & \bf\rotatebox{90}{train} & \bf\rotatebox{90}{motor} & \bf\rotatebox{90}{bike} & \bf mIoU & \bf Avg. \\ \hline

    \multirow{2}*{\emph{Individual}~\cite{vu2018advent}} & M & 81.1 & 18.6 & 74.8 & 23.9 & 28.9 & 30.3 & 35.7 & 33.7 & 78.4 & 40.7 & 93.3 & 49.5 & \textbf{42.3} & 80.4 & 35.1 & 34.2 & \textbf{17.8} & \textbf{41.8} & 36.1 & 46.1 & \multirow{2}*{44.9} \\  
    ~ & I & \textbf{94.1} & 24.4 & 66.1 & 31.3 & \textbf{22.0} & 25.4 & 9.3 & 26.7 & 80.0 & 31.4 & 93.5 & 48.7 & 43.8 & 71.4 & 49.4 & 28.5 & \textbf{0} & 48.7 & \textbf{34.3} & 43.6 & ~ \\ \hline
    
    \multirow{2}*{\emph{MTKT} \cite{saporta2021mtaf}} & M & \textbf{85.4} & \textbf{42.8} & 78.6 & 28.9 & \textbf{32.8} & 31.0 & 32.8 & 35.4 & \textbf{79.8} & \textbf{45.0} & \textbf{95.1} & 54.1 & 34.3 & 82.5 & 36.2 & 34.1 & 9.5 & 40.6 & \textbf{37.6} & 48.2 & \multirow{2}*{45.4} \\  
    ~ & I & 82.1 & 14.1 & 56.4 & 31.4 & 21.3 & 28.6 & 12.5 & 43.0 & \textbf{81.0} & 26.9 & \textbf{93.6} & 35.8 & 46.9 & 76.2 & 56.9 & 39.5 & \textbf{0} & 50.5 & 13.4 & 42.6 & ~\\ \hline
    
    \multirow{2}*{\emph{\ours} (Ours)} & M & 79.4 & 35.3 & \textbf{81.0} & \textbf{34.6} & 30.9 & \textbf{37.8} & \textbf{43.7} & \textbf{52.7} & 74.1 & \textbf{45.0} & 93.4 & \textbf{63.7} & \textbf{43.2} & \textbf{84.9} & \textbf{48.2} & \textbf{51.3} & 5.3 & 39.8 & 36.6 & \textbf{51.6} & \multirow{2}*{\textbf{50.6}} \\  
    ~ & I & 87.1 & \textbf{30.1} & \textbf{66.3} & \textbf{34.7} & 21.8 & \textbf{34.5} & \textbf{18.9} & \textbf{66.0} & 80.6 & \textbf{41.7} & 91.3 & \textbf{52.8} & \textbf{55.8} & \textbf{83.7} & \textbf{58.4} & \textbf{48.0} & \textbf{0} & \textbf{55.6} & 13.2 & \textbf{49.5} & ~\\ \hline

    \hline
    \noalign{\smallskip}
    \multicolumn{23}{c}{\bf GTA5 $\rightarrow$ Cityscapes + Mapillary + IDD}\\
    \noalign{\smallskip}
    \hline
    \bf Method & \bf \rotatebox{90}{Target} & \bf\rotatebox{90}{road} & \bf\rotatebox{90}{sidewalk} & \bf\rotatebox{90}{building} & \bf\rotatebox{90}{walk} & \bf\rotatebox{90}{fence} & \bf\rotatebox{90}{pole} & \bf\rotatebox{90}{light} & \bf\rotatebox{90}{sign} & \bf\rotatebox{90}{veg} & \bf\rotatebox{90}{terrain} & \bf\rotatebox{90}{sky} & \bf\rotatebox{90}{person} & \bf\rotatebox{90}{rider} & \bf\rotatebox{90}{car} & \bf\rotatebox{90}{truck} & \bf\rotatebox{90}{bus} & \bf\rotatebox{90}{train} & \bf\rotatebox{90}{motor} & \bf\rotatebox{90}{bike} & \bf mIoU & \bf Avg. \\ \hline
    
    \multirow{3}*{\emph{Individual}~\cite{vu2018advent}} & C & \textbf{88.8} & 23.8 & \textbf{81.5} & 27.7 & 27.3 & 31.7 & 33.2 & 22.9 & 83.1 & 27.0 & 76.4 & 58.5 & \textbf{28.9} & 84.3 & 30.0 & 36.8 & 0.3 & 27.7 & \textbf{33.1} & 43.3 & \multirow{3}*{43.3} \\ 
    ~ & M & 81.1 & 18.6 & 74.8 & 23.9 & 28.9 & 30.3 & 35.7 & 33.7 & \textbf{78.4} & 40.7 & 93.3 & 49.5 & 42.3 & 80.4 & 35.1 & 34.2 & 17.8 & \textbf{41.8} & \textbf{36.1} & 46.1 & ~ \\
    ~ & I & \textbf{94.1} & 24.4 & \textbf{66.1} & 31.3 & \textbf{22.0} & 25.4 & 9.3 & 26.7 & 80.0 & \textbf{31.4} & \textbf{93.5} & 48.7 & 43.8 & 71.4 & 49.4 & 28.5 & \textbf{0} & 48.7 & \textbf{34.3} & 43.6 & ~ \\ \hline
    
    \multirow{3}*{\emph{MTKT} \cite{saporta2021mtaf}} & C & 85.9 & 33.7 & 81.2 & 30.2 & 20.0 & 31.3 & 32.7 & 17.5 & 84.1 & 33.2 & 80.8 & 56.1 & 16.8 & 83.2 & 26.0 & 39.2 & \textbf{10.9} & 24.4 & 13.7 & 42.2  & \multirow{3}*{43.9} \\  
    ~ & M & \textbf{85.9} & 42.1 & 76.1 & 29.1 & 28.7 & 30.3 & 35.1 & 34.4 & 76.6 & 43.1 & \textbf{93.7} & 55.2 & 31.1 & 82.7 & 37.8 & 31.8 & \textbf{20.8} & 37.8 & 17.3 & 46.8 & ~ \\
    ~ & I & 83.1 & 16.9 & 55.7 & \textbf{34.7} & 21.3 & 27.6 & 8.0 & 48.6 & 77.7 & 26.9 & 91.9 & 36.7 & 46.2 & 74.2 & 56.1 & 41.3 & \textbf{0} & 48.7 & 15.7 & 42.7 & ~ \\ \hline
    
    \multirow{3}*{\emph{\ours} (Ours)} & C & 88.4 & \textbf{43.0} & 80.0 & \textbf{30.9} & \textbf{29.4} & \textbf{37.6} & \textbf{36.9} & \textbf{42.1} & \textbf{86.2} & \textbf{40.9} & \textbf{81.5} & \textbf{60.1} & 15.4 & \textbf{85.5} & \textbf{33.5} & \textbf{44.5} & 4.6 & \textbf{30.7} & 26.7 & 47.2  & \multirow{3}*{\textbf{49.1}} \\  
    ~ & M & 82.8 & \textbf{44.8} & \textbf{79.5} & \textbf{32.3} & \textbf{37.9} & \textbf{38.3} & \textbf{38.2} & \textbf{52.4} & 76.0 & \textbf{45.5} & 92.9 & \textbf{65.2} & \textbf{39.2} & \textbf{85.8} & \textbf{51.0} & \textbf{43.1} & 6.2 & 38.2 & 27.3 & \textbf{51.4} & ~ \\
    ~ & I & 86.9 & \textbf{29.0} & 64.1 & 31.2 & 20.2 & \textbf{36.7} & \textbf{14.8} & \textbf{51.9} & \textbf{82.3} & \textbf{48.2} & 92.7 & \textbf{51.8} & \textbf{53.6} & \textbf{83.8} & \textbf{60.7} & \textbf{46.6} & \textbf{0} & \textbf{50.5} & 20.6 & \textbf{48.7} & ~ \\ \hline
    
    \end{tabular}
}
\caption{The detailed class-wise comparison of \emph{\ours} in the \emph{19-class} setting with the existing state-of-the-art methods. In all the experiments, GTA5 is considered as the source domain and the various combinations of the other benchmarks are considered as the target domains. In all the configurations our \ours clearly outperforms the existing baselines by a clear margin}
\label{table:detail19}
\end{center}
\end{table*}
\setlength{\tabcolsep}{1.4pt}

\subsection{Synthetic to Real scenario: summary of all the Settings.}
In the main paper, we report in Tab.~\ref{table:sotaSummary} the average mIoU considering all the possible target configurations on the \emph{19-classes} Benchmark in the \emph{Synthetic to Real} scenario. We now report the results on the \emph{7-classes} Benchmark in Tab.~\ref{table:sotaSummary7}. In short, we observe that \emph{\ours} obtains performance on par with \emph{ADAS}~\cite{lee2022adas}. \emph{\ours} obtains the pest average performance in three configurations over four. These experiments demonstrate again the robustness of our approach.

\setlength{\tabcolsep}{1.4pt}
\begin{table}[!h]
\begin{center}

\resizebox{0.4\textwidth}{!}{\begin{tabular}{c c c|c| c c c |c}

 \multicolumn{8}{c}{\bf \emph{7-classes} Benchmark}\\
\hline
 \multicolumn{3}{c|}{\bf Target} &  \multirow{2}*{\bf method} & \multicolumn{3}{c|}{\bf mIoU} & \bf mIoU\\ 
 C & I & M & ~ & C & I & M & Avg.\\ \hline

\multirow{3}*{$\surd$} & \multirow{3}*{$\surd$} & \multirow{3}*{-} & \emph{MTKT}~\cite{saporta2021mtaf} & 70.4 & 65.9 & - & 68.2 \\ 
 ~ & ~ & ~ & \emph{ADAS}~\cite{lee2022adas}($1024\times512$) & \bf 75.4 & 66.9 & - &  71.2 \\
 ~ & ~ & ~ & \ours (Ours) & 72.6 & \textbf{70.0} & - &  \bf 71.3 \\ \hline
  \multirow{3}*{$\surd$} & \multirow{3}*{-} & \multirow{3}*{$\surd$} & \emph{MTKT}~\cite{saporta2021mtaf} & 71.1 & - & 70.8 & 71.0 \\ 
 ~ & ~ & ~ & \emph{ADAS}~\cite{lee2022adas}($1024\times512$) & \bf 75.3 & - & \bf 72.6 &  \bf 73.9 \\
 ~ & ~ & ~ & \ours (Ours) & 72.3 & - & 72.3 & 72.3 \\ \hline
  \multirow{3}*{-} & \multirow{3}*{$\surd$} & \multirow{3}*{$\surd$} & \emph{MTKT}~\cite{saporta2021mtaf} & - & 65.9 & 70.7 & 68.3 \\ 
 ~ & ~ & ~ & \emph{ADAS}~\cite{lee2022adas}($1024\times512$) & - & - & - &  - \\
 ~ & ~ & ~ & \ours (Ours) & - & \bf 68.7 & \bf 72.4 & \bf 70.6 \\ \hline
  \multirow{3}*{$\surd$} & \multirow{3}*{$\surd$} & \multirow{3}*{$\surd$} & \emph{MTKT}~\cite{saporta2021mtaf} & 70.4 & 65.9 & 71.1 & 69.1 \\ 
 ~ & ~ & ~ & \emph{ADAS}~\cite{lee2022adas}($1024\times512$) & \bf 74.9 & 66.7 & \bf 72.2 &  71.3 \\
 ~ & ~ & ~ & \ours (Ours) & \textbf{72.6} & 70.3 & \textbf{72.2} & \bf 71.7 \\ \hline

\end{tabular}}
\caption{Summary of performances obtained on the \emph{7-classes} Benchmark with different dataset configurations. Cityscapes, IDD and Mapillary are referred to as C, I and M respectively. We report the mIoU averaged over the target domains.}
\label{table:sotaSummary7}
\end{center}
\end{table}
\setlength{\tabcolsep}{1.4pt}

\subsection{Detailed Results of the Real to Real Settings}
Here we show the comparison with \emph{MTKT} in all the Real to Real settings on the \emph{7-class} benchmark. We observe from the Tab.~\ref{table:R2R} that \emph{\ours} can clearly outperform \emph{MTKT} in all the real to real configurations. This again proves the versatility of \emph{\ours} as it can yield better performance when trained on both synthetic and real source domains.

\setlength{\tabcolsep}{1.4pt}
\begin{table*}[!ht]
\begin{center}

\resizebox{0.8\textwidth}{!}{
    \begin{tabular}{c|c|c c c c c c c |c|c}
    \hline
    \noalign{\smallskip}
    \multicolumn{11}{c}{\bf Cityscapes $\rightarrow$ Mapillary + IDD}\\ 
    \noalign{\smallskip}
    \hline
    \bf Method & \bf Target & \bf flat & \bf constr & \bf object & \bf nature & \bf sky & \bf human & \bf vehicle & \bf mIoU & \bf Avg. \\ \hline
    
    \multirow{2}*{\emph{MTKT}\cite{saporta2021mtaf}} & M & 88.3 & 70.4 & 31.6 & 75.9 & \bf 94.4 & 50.9 & 77.0 & 69.8 & \multirow{2}*{69.0} \\  
    ~ & I & 93.6 & 54.9 & 18.6 & 84.0 & \textbf{94.5} & 53.4 & 79.2 & 68.3 & ~ \\ \hline

    \multirow{2}*{\emph{\ours} (Ours)} & M & \textbf{90.2} & \textbf{73.4} & \textbf{37.2} & \textbf{78.8} & 92.3 & \textbf{59.2} & \textbf{84.1} & \textbf{73.6} & \multirow{2}*{\textbf{72.6}} \\  
    ~ & I & \textbf{95.1} & \textbf{58.0} & \textbf{26.6} & \textbf{85.4} & 93.0 & \textbf{59.0} & \textbf{83.9} & \textbf{71.6} & ~ \\ \hline

    \hline
    \noalign{\smallskip}
    \multicolumn{11}{c}{\bf Mapillary $\rightarrow$ Cityscapes + IDD}\\ 
    \noalign{\smallskip}
    \hline
    \bf Method & \bf Target & \bf flat & \bf constr & \bf object & \bf nature & \bf sky & \bf human & \bf vehicle & \bf mIoU & \bf Avg. \\ \hline
    
    \multirow{2}*{\emph{MTKT}\cite{saporta2021mtaf}} & C & 94.7 & 81.9 & 35.6 & 83.0 & 84.7 & 57.0 & 83.9 & 74.4 & \multirow{2}*{72.7} \\  
    ~ & I & 95.2 & 61.6 & 24.6 & 85.4 & 94.3 & 55.7 & 81.1 & 71.1 & ~ \\ \hline

    \multirow{2}*{\emph{\ours} (Ours)} & C & \textbf{95.6} & \textbf{84.4} & \textbf{36.7} & \textbf{83.9} & \textbf{88.2} & \textbf{58.2} & \textbf{85.8} & \textbf{76.1} & \multirow{2}*{\textbf{74.8}} \\  
    ~ & I & \textbf{95.5} & \textbf{64.6} & \textbf{31.1} & \textbf{85.8} & \textbf{94.6} & \textbf{58.2} & \textbf{84.7} & \textbf{73.5} & ~ \\ \hline

    \hline
    \noalign{\smallskip}
    \multicolumn{11}{c}{\bf IDD $\rightarrow$ Cityscapes + Mapillary}\\ 
    \noalign{\smallskip}
    \hline
    \bf Method & \bf Target & \bf flat & \bf constr & \bf object & \bf nature & \bf sky & \bf human & \bf vehicle & \bf mIoU & \bf Avg. \\ \hline
    
    \multirow{2}*{\emph{MTKT}\cite{saporta2021mtaf}} & C & \textbf{96.7} & 82.8 & 31.0 & \textbf{84.7} & \textbf{89.8} & \textbf{60.2} & 85.1 & 75.8 & \multirow{2}*{73.9} \\  
    ~ & M & 90.4 & 71.2 & 33.8 & 79.1 & 95.8 & 55.3 & 79.0 & 72.1 & ~ \\ \hline

    \multirow{2}*{\emph{\ours} (Ours)} & C & 96.5 & \textbf{84.3} & \textbf{33.6} & \textbf{84.7} & 89.1 & 58.3 & \textbf{85.8} & \textbf{76.0} & \multirow{2}*{\textbf{75.7}} \\  
    ~ & M & \textbf{91.0} & \textbf{76.3} & \textbf{39.7} & \textbf{82.2} & \textbf{96.0} & \textbf{59.0} & \textbf{83.1} & \textbf{75.3} & ~ \\ \hline
    
    \end{tabular}
}
\caption{The detailed comparison of \ours with the state-of-the-art on the \emph{7-classes} benchmark in all Real to Real scenarios. \ours clearly outperforms MTKT in all the real to real configuration. }
\label{table:R2R}
\end{center}
\end{table*}
\setlength{\tabcolsep}{1.4pt}

\subsection{Comparison with other MTDA methods.}


In this section, we compare our methods with other MTDA methods in the literature that have been proposed for object recognition. Following \emph{CCL}~\cite{isobe2021multi}, we report the numbers of \emph{\ours} on the {\em 19-class} benchmark in the Tab.~\ref{table:comparison}. Note that only \emph{CCL}~\cite{isobe2021multi} and \emph{\ours} are specifically designed for semantic segmentation. The baselines in the MTDA setting \cite{gholami2020unsupervised,nguyen2021unsupervised} that are designed for object recognition perform sub-optimally with respect \emph{CCL}~\cite{isobe2021multi}. However, \emph{\ours} surpasses \emph{CCL}~\cite{isobe2021multi} by a non-trivial margin, validating the importance of data driven image stylization for the MTDA in semantic segmentation.

\setlength{\tabcolsep}{1.4pt}
\begin{table}[!h]
\begin{center}
\resizebox{0.44\textwidth}{!}{
\begin{tabular}{c|c|cc|c}

\hline
\noalign{\smallskip}
\multicolumn{5}{c}{\bf GTA5 $\rightarrow$ Cityscapes + IDD}\\ 
\noalign{\smallskip}
\hline
\multirow{2}*{\bf Setting} & \multirow{2}*{\bf Method} &  \multicolumn{2}{c|}{\bf mIoU} & \multirow{2}*{\bf Avg.}\\ 
~ & ~ & C & I & ~ \\ 
\hline

DG & \emph{Yue et al.}~\cite{yue2019domain} & 42.1 & 42.8 & 42.5\\ \hline

\multirow{4}*{MTDA} & \emph{MTDA-ITA}~\cite{gholami2020unsupervised} & 40.3 & 41.2 & 40.8 \\ 
~  & \emph{MT-MTDA}~\cite{nguyen2021unsupervised} & 43.2 & 44.0 & 43.6 \\ 
~  & \emph{CCL}~\cite{isobe2021multi} & 45.0 & 46.0 & 45.5 \\ 
~  & \emph{\ours} (Ours) & \bf 47.1 & \bf 49.3  & \bf 48.2 \\
\hline
\end{tabular}}
\caption{The quantitative comparison of our \ours with different MTDA methods on the \emph{19-class} benchmark for the GTA5 $\rightarrow$ Cityscapes + IDD configuration. \ours outperforms the considered MTDA baselines that are designed for either object recognition or semantic segmentation. DG stands for domain generalization setting and a method designed for such a setting is also under performed by \ours}
\label{table:comparison}

\end{center}
\end{table}
\setlength{\tabcolsep}{1.4pt}






\subsection{Direct Transfer to Unseen Domains}

Similar to \cite{saporta2021mtaf}, we directly test our adapted model on a new (or \textit{unseen}) target domain to evaluate the generalization ability of our model. This setting is often referred to as \textit{open-compound} domain adaptation in the literature. In the Tab.~\ref{table:transfer}, we report the comparison of the generalization ability with other methods on \emph{7-class} benchmark. We can observe that among considered MTDA baselines, our \emph{\ours} has the best generalization ability. This hints at the fact that our proposed cooperative self-training realized with feature stylization can induce better generalizability.


\setlength{\tabcolsep}{1.4pt}
\begin{table*}[!t]
\begin{center}

\resizebox{0.8\textwidth}{!}{
    \begin{tabular}{c|c|c|c c c c c c c |c}

    \hline
    \noalign{\smallskip}
    \multicolumn{11}{c}{\bf Direct Transfer to an Unseen Target Domain}\\ 
    \noalign{\smallskip}
    \hline
    \bf Setup & \bf Method & \bf Test & \bf flat & \bf constr & \bf object & \bf nature & \bf sky & \bf human & \bf vehicle & \bf mIoU  \\ \hline
    
    \multirow{4}*{G $\rightarrow$ C + I} & \emph{Data Comb.}~\cite{vu2018advent} & \multirow{4}*{M} & 88.4 & 71.0 & 31.0 & 72.4 & 92.0 & 37.4 & 74.7 & 66.7 \\  
    ~ & \emph{Multi-Dis}\cite{saporta2021mtaf} & ~ & 89.2 & 72.1 & 21.7 & 73.8 & \bf 94.0 & 34.8 & 75.9 & 65.9 \\
    ~ & \emph{MTKT}\cite{saporta2021mtaf} & ~ & 89.8 & \bf 74.0 & 30.4 & 74.1 & 93.6 & 52.6 & 79.4 & 70.6 \\
    ~ & \emph{\ours} (Ours) & ~ & \bf 91.6 & 73.9 & \bf 34.8 & \bf 77.8 & 93.0 & \bf 57.7 & \bf 81.9 & \bf 72.9 \\
    \hline
    
    \multirow{4}*{G $\rightarrow$ C + M} & \emph{Data Comb.}\cite{vu2018advent} & \multirow{4}*{I} & 91.6 & 54.7 & 13.9 & 76.5 & 90.9 & 48.3 & 77.5 & 64.8 \\  
    ~ & \emph{Multi-Dis}\cite{saporta2021mtaf} & ~ & 91.2 & 54.6 & 12.9 & 77.7 & \bf 92.5 & 50.3 & 78.6 & 65.4 \\
    ~ & \emph{MTKT}\cite{saporta2021mtaf} & ~ & 91.5 & 56.1 & 12.3 & 76.1 & 90.9 & 51.4 & 79.2 & 65.4 \\
    ~ & \emph{\ours} (Ours) & ~ & \bf 93.2 & \bf 59.7 & \bf 17.1 & \bf 80.1 & 91.0 & \bf 51.7 & \bf 81.2 & \bf 67.7 \\
    \hline
    
    
    \end{tabular}
}
\caption{The quantitative comparison for direct transfer to new (or \textit{unseen}) domains in \emph{7-class} benchmark. GTA5, Cityscapes, Mapillary and IDD are referred to as G, C, M and I, respectively. The \textbf{Test} column denotes the unseen target domain where the models have been evaluated}
\label{table:transfer}
\end{center}
\end{table*}
\setlength{\tabcolsep}{1.4pt}







\end{document}